\documentclass[11pt]{article}

\usepackage[preprint]{acl}

\usepackage{times}
\usepackage{latexsym}
\usepackage{algorithm}
\usepackage{algpseudocode}
\usepackage[T1]{fontenc}

\usepackage[utf8]{inputenc}
\usepackage{subcaption}
\usepackage{microtype}
\usepackage{booktabs}
\usepackage{tabularx}
\usepackage{inconsolata}

\usepackage{graphicx}
\usepackage{amsmath}
\usepackage{amssymb}
\usepackage[table]{xcolor}

\usepackage{xcolor}
\usepackage{amsmath}
\usepackage{booktabs}
\usepackage{xspace}
\usepackage{multirow}
\newcommand{\method}{\textbf{ReRULE}\xspace}
\usepackage{booktabs}
\usepackage{multirow}
\usepackage{listings}
\usepackage[table]{xcolor}
\usepackage{arydshln}
\usepackage{stfloats}
\usepackage{alltt}
\usepackage{placeins}
\usepackage{tcolorbox}

\definecolor{pinkpurple}{RGB}{218, 112, 214}

\usepackage{CJKutf8}
\title{Replay What Matters: Off-Policy Replay \\for Efficient LLM Reinforcement Unlearning}

\author{
Zirui Pang\textsuperscript{1}, Chenlong Zhang\textsuperscript{2}, Haosheng Tan\textsuperscript{3}, Zhuoran Jin\textsuperscript{2}, Jiaheng Wei\textsuperscript{1}, Zixin Zhong\textsuperscript{1}\thanks{Corresponding author: zixinzhong@hkust-gz.edu.cn} \\
\textsuperscript{1}The Hong Kong University of Science and Technology (Guangzhou),\\ 
\textsuperscript{2}Institute of Automation, Chinese Academy of Sciences,
\textsuperscript{3}University of Glasgow
}

\begin{document}
\maketitle
\begin{abstract}
LLM unlearning has emerged as a cost-effective alternative to full retraining for removing hazardous knowledge from pretrained models while preserving general utility.
Recent RL-based methods such as RULE reformulate unlearning as learning a refusal behavior, but their on-policy optimization repeatedly samples from the same forget and retain/boundary prompts throughout training.
We identify a critical inefficiency in this process: easy cases quickly converge and provide little useful gradient signal, while hard cases near the forget/retain boundary continue to produce low-reward rollouts that are discarded after a single use.
To address this issue, we propose \method, an off-policy replay enhancement for reinforcement unlearning.
\method stores low-reward hard-case rollout groups in a replay buffer during early GRPO training and reuses them in later stages through importance-sampled off-policy updates, redirecting computation toward boundary cases that  require further learning.
Theoretically, we show that \method yields a tighter hard-case convergence bound than pure on-policy RULE.
Empirically, \method improves MUSE-Books Retain Quality from 46.3 to 56.2 while adding only 5--11\% training time across benchmarks.
Its limited improvement on the simpler TOFU setting further supports the intended conditional behavior: replay is most beneficial when the hard/easy disparity is pronounced.
\end{abstract}

\section{Introduction}
Recent advances in large language model (LLM) pretraining have significantly improved model capabilities \citep{yang2025qwen3technicalreport,team2023gemini,guo2025deepseek,touvron2023llama,zhang2022opt,li2023textbooks,pu2026cameoconditionalqualityawaremultiagent,li2026ipbenchbenchmarkimageprotection,li2026recognition}. However, because these models are trained on large-scale internet corpora \citep{yang2025entp,deng2025lm,wu2026lifeside,liu2025selectmix}, they may acquire copyrighted content \citep{karamolegkou2023copyright,grynbaum2023times,chu2024protect,zhang2024unlearncanvas,zhang2024generate,ji2024reversing}, sensitive private information \citep{staab2024beyond,mireshghallah2024can,das2025security,di2025adversarial,zheng2026offsidebenchmarkingunlearningmisinformation,jia2024wagle,fan2025towards}, or other hazardous knowledge\citep{sun2026stegobattlefieldevaluatingimage,jia2026evianexplainablevisualinstructiontuning,wang2025invariancemakesllmunlearning,reisizadeh2026blur,zhuang2025seuf}. Removing such data and retraining from scratch is prohibitively expensive, which hence motivates LLM unlearning \citep{yao2024large}: methods that remove undesired knowledge while preserving the model's general utility.

Reinforcement learning (RL) offers a natural route for unlearning because it can directly optimize refusal behavior. 
RULE \citep{zhang2025rulereinforcementunlearningachieves} follows this idea by reformulating unlearning as teaching the model to refuse queries related to the forget set while preserving responses on retain and boundary queries. Unlike SFT-based unlearning methods that process each sample once, RL-based methods such as RULE repeatedly sample rollouts from the same training prompts. Such repetition is useful for exploration in the beginning of training phase, but it becomes inefficient once many prompts have already converged.

We observe a pronounced \textbf{hard--easy disparity} in the unlearning data. As shown in Figure~\ref{fig:sampling}, some prompts rapidly achieve high and stable rewards, whereas others remain low-reward throughout training and require substantially more optimization effort. These hard cases often lie near the semantic boundary between forgetting target knowledge and preserving useful retain behavior. In late-stage on-policy GRPO\citep{guo2025deepseek}, repeatedly sampling already-converged easy cases provides little informative gradient signal. Meanwhile, hard cases are also under-utilized: their low-reward rollouts often lead to ineffective updates and are discarded after a single use, despite containing valuable information about the model's failure modes.

Motivated by this training dynamic, we propose Replay-enhanced Reinforcement UnLEarning(\method), an off-policy replay enhancement mechanism. During early Stage 2 GRPO training, \method stores rollout groups whose mean reward falls below a threshold into a replay buffer. Once the buffer contains sufficient hard cases and training has completed a dataset-specific warm-up step, \method performs hybrid training: it keeps the standard on-policy GRPO update while adding an importance-sampled off-policy update over replayed hard cases. This redirects the attention of gradient computation from converged easy cases to boundary cases that still require further learning.

Empirically, this targeted replay improves the setting where the hard/easy disparity is highly pronounced. On MUSE-Books, \method improves Retain Quality from 46.3 to 56.2 while preserving strong forgetting performance. The additional training cost remains modest, increasing total training time by only 5--11\% across RWKU, MUSE, and TOFU. On TOFU, where the forget01 setting is smaller and hence easier, \method brings limited improvement over RULE; this conditional behavior supports the claim that replay is most valuable when persistent hard cases are present.

In summary, our contributions are threefold:

\begin{itemize}
    \item We identify a training inefficiency in RL-based LLM unlearning: on-policy GRPO spends late-stage computation on converged easy cases while hard cases near the forget/retain boundary remain under-optimized.
    
    \item We propose \method, an off-policy hard-case replay method that stores low-reward rollout groups and reuses them through importance-sampled updates, thereby improving the training efficiency of reinforcement unlearning. We further theoretically show that \method tightens the hard-case convergence bound.
    
    \item We demonstrate the effectiveness of replay-based rehearsal for unlearning on three benchmarks: \method achieves better performance than the baselines across all benchmarks. In particular, it improves the Retain Quality on MUSE-Books by 9.9 absolute points over RULE while introducing only modest additional training overhead. Moreover, when the hard-case storage mechanism is replaced with random storage, the performance of \method degrades to a level comparable to RULE.

\end{itemize}

\section{Related Work}
In this section, we discuss existing LLM unlearning methods and summarize their key characteristics.
\paragraph{Generation-based LLM Unlearning methods.}Generation-based methods refer to training-free approaches \citep{pang2025labelsmoothingimprovesgradient} that do not modify model parameters or alter the model’s internal representations. Instead, they steer the model’s output distribution away from harmful regions through prompt manipulation or output-level adjustments \citep{gao2024large,thaker2024guardrail,muresanu2024fast,pawelczyk2023context}. Guard \citep{deng2025guardgenerationtimellmunlearning} trains a classifier to determine whether the outputs of a large language model contain hazardous information related to the forget data. ECO Prompt \citep{liu2024large} also trains a classifier, but applies lightweight perturbations to steer the outputs toward safe responses. In contrast, Soft Prompt Unlearning \citep{bhaila2025soft} avoids hazardous outputs by appending learnable soft prompts to the inputs.

\paragraph{Fine-tuning-based LLM Unlearning methods.}Fine-tuning-based methods refer to approaches that achieve unlearning by updating model parameters through training \citep{pang2025labelsmoothingimprovesgradient}. The Gradient Ascent (GA) \citep{bourtoule2021machine} method inverts the loss used in supervised fine-tuning, effectively turning gradient descent on target knowledge into gradient ascent, thereby inducing forgetting. Gradient Difference (GD) \citep{wang2023kga} further extends GA by incorporating retain data into the optimization process. NPO \citep{zhang2024negative} introduces a reference model to impose a lower-bound constraint on the optimization, thereby mitigating the collapse risk of GA. SimNPO \citep{fan2025simplicityprevailsrethinkingnegative} removes the reference model and provides a simplified variant of NPO. Similarly, the lower-bound regularization principle in NPO is also adopted in DPO and KTO \citep{ethayarajh2024kto}. FLAT \citep{wang2024llm} achieves unlearning by maximizing the f-divergence between the available template answers and the forget targets. SGA \citep{pang2025labelsmoothingimprovesgradient} introduces normal data that are similar to the forget data, enabling the model to continue learn from normal data while performing unlearning, and thus alleviating the instability caused by gradient ascent. RULE \citep{zhang2025rulereinforcementunlearningachieves} reformulates unlearning as a rewardable refusal behavior and leverages reinforcement learning to achieve forgetting. 

Motivated by the under-optimization of hard cases in the stage-two training of RULE, we propose \method, which improves both the performance on hard cases and the training efficiency of LLM reinforcement unlearning.

\section{\method: Replay Enhancement for Reinforcement Unlearning}
In this section, we first summarize the prior work RULE and explain our motivation. We then introduce the overall pipeline of \method.
\subsection{RULE: Reinforcement UnLEarning}
In the LLM unlearning literature, conventional methods typically optimize a joint objective over the forget and retain sets:
\begin{equation}
\begin{aligned}
\min_{\theta} \quad
& \underbrace{\mathbb{E}_{(x_f,y_f)\in \mathcal{D}_f}
[\ell_f(y_f \mid x_f;\theta)]}_{\text{forget}} \\
& + \lambda
\underbrace{\mathbb{E}_{(x_r,y_r)\in \mathcal{D}_r}
[\ell_r(y_r \mid x_r;\theta)]}_{\text{retain}} .
\end{aligned}
\end{equation}
Here, $\theta$ denotes the model parameters, when $\mathcal{D}_f$ and $\mathcal{D}_r$ denote the forget and retain sets, respectively, and $x$ and $y$ represent the input query and its corresponding target response~\citep{yao2024large}.

However, direct optimization of such objectives can substantially degrade model utility. To mitigate this issue, RULE~\citep{zhang2025rulereinforcementunlearningachieves} reformulates unlearning as learning to refuse rather than directly forgetting. It trains the model with reinforcement learning techniques to behave differently on the two sides of a refusal boundary. The specific procedure is as follows:

\paragraph{Stage 1: Rejection Steering.}
RULE first applies supervised learning on the forget set $\mathcal{D}_f$ to warm-start the model toward refusal behavior, enabling the model to learn how to refuse.
The construction details of the refusal supervision data are provided in Appendix~\ref{app:data_generation}. 
The optimization objective is:
\begin{equation*}
\theta_{\text{rej}} = \arg\max_{\theta} \;
\mathbb{E}_{(x, y^*) \sim \mathcal{D}_f}
\left[ \log \pi_{\theta}(y^* \mid x) \right],
\end{equation*}
where $y^*$ denotes the corresponding refusal response.
\paragraph{Stage 2: Refusal Boundary Optimization.}
After the refusal SFT stage, RULE further refines the refusal boundary through on-policy RL, enabling the model to distinguish between forget queries and retain queries on top of the learned refusal behavior.
Specifically, the policy is optimized on the union of the forget set $\mathcal{D}_f$ and the constructed boundary retain set $\widetilde{\mathcal{D}}_r$. 
The construction details of $\widetilde{\mathcal{D}}_r$ are provided in Appendix~\ref{app:data_generation}.

Specifically, RULE optimizes the policy using on-policy RL methods, such as PPO~\citep{schulman2017proximalpolicyoptimizationalgorithms} and GRPO~\citep{shao2024deepseekmathpushinglimitsmathematical}, with a KL regularization term anchoring the policy to $\pi_{\text{rej}}$:
\begin{equation}
\max_{\theta} \;
\mathbb{E}_{x,\, y \sim \pi_\theta}
\left[ r(x,y) \right]
- \beta \, D_{\mathrm{KL}} \left[ \pi_\theta \,\|\, \pi_{\text{rej}} \right].
\end{equation}

The reward function $r(x,y)$ in RULE is defined as: {\scriptsize
\begin{equation*}
\begin{cases}
\lambda_f \cdot \mathbb{I}[y \in \mathcal{P}_{\text{refuse}}]
+ (1 - \lambda_f) \cdot \mathbb{I}[k(x) \subset y],
\quad x \in \mathcal{D}_f, \\[4pt]
\lambda_r \cdot \mathbb{I}[y \notin \mathcal{P}_{\text{refuse}}]
+ (1 - \lambda_r) \cdot \mathbb{I}[\text{ROUGE-L}(y, y_{\text{gold}}) > \gamma],
\quad x\in \widetilde{\mathcal{D}}_r.
\end{cases}
\end{equation*}
}

\subsection{Hard Cases in RULE Stage 2 Training.}
\label{method}
\begin{figure}[t]
    \centering

    \begin{subfigure}[b]{0.7\linewidth}
        \centering
        \includegraphics[width=\linewidth]{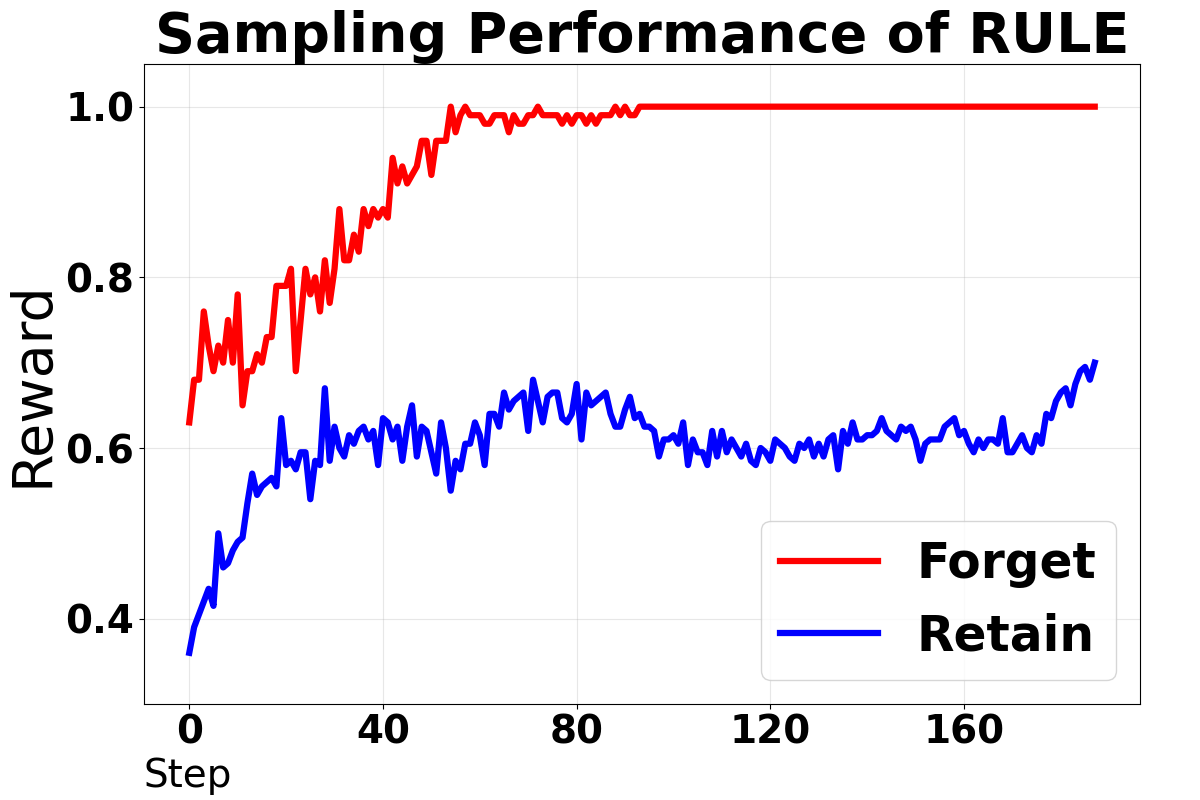}
        \caption{Reward by prompt type}
        \label{fig:sampling_a}
    \end{subfigure}

    \vspace{0.3cm}

    \begin{subfigure}[b]{1\linewidth}
        \centering
        \includegraphics[width=\linewidth]{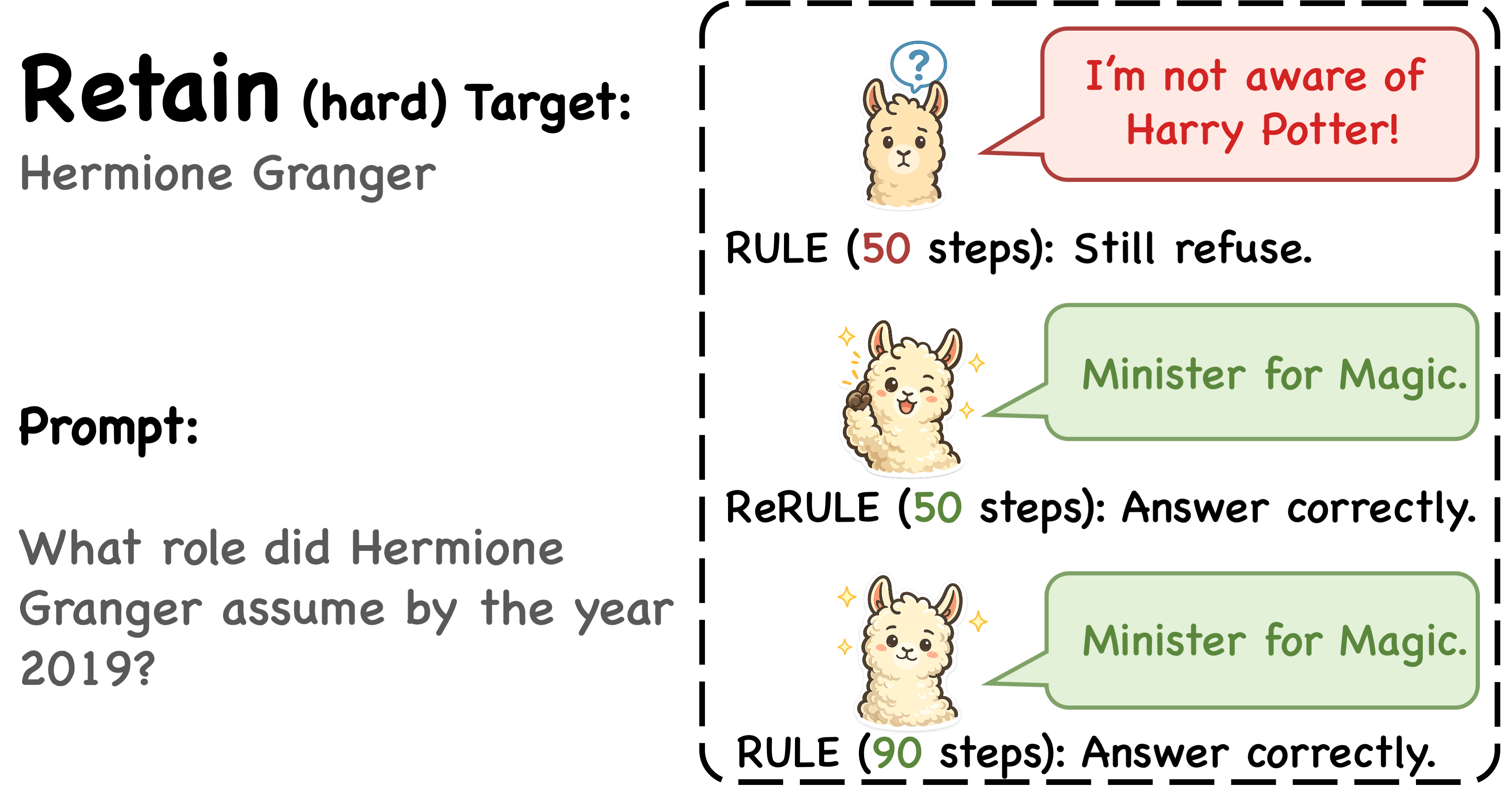}
        \caption{Different answers for the same step}
        \label{fig:sampling_b}
    \end{subfigure}

    \caption{(a) Validation-set rewards during Stage 2 training of RULE on the MUSE-Books task. Questions are categorized into Forget and Retain groups. We observe that there consistently exist questions with relatively poor sampling performance throughout training. (b) For hard cases, RULE requires nearly twice as many training steps to achieve satisfactory performance, whereas \method can save these additional steps.}
    \label{fig:sampling}
    \vspace{-0.2cm}
\end{figure}

In Stage 2 of RULE, we observe a critical issue: the model exhibits significantly different sampling behaviors across questions of different types and topics. As illustrated in figure~\ref{fig:sampling}, the model shows substantial variance in sampling performance on the MUSE-Books task. Based on these performance disparities, we can naturally divide the training samples into two parts: Hard cases and Easy cases.

Meanwhile, the gradient of on-policy GRPO \citep{shao2024deepseekmathpushinglimitsmathematical} can be written as:
\begin{equation}
\begin{aligned}
\nabla_\theta J_{\text{on}}(\theta)
&=
\mathbb{E}_{x\sim p(x),\, y\sim\pi_\theta(\cdot|x)}
\Bigl[
\\
&\quad
\underbrace{
\nabla_\theta \log \pi_\theta(y|x)
\cdot
\hat A(x,y)
}_{g(x,y)}
\Bigr]
\end{aligned},
\label{eq:on_policy_gradient}
\end{equation}
where \(x\) denotes the sampled prompt, \(y\) denotes the rollout generated by the current policy \(\pi_\theta\), and \(\hat A(x,y)\) represents the group-normalized GRPO advantage.

Given a predefined threshold $\tau$, we partition the prompt space into \textbf{Easy} and \textbf{Hard} regions based on the group mean reward and $\tau$.

The \textbf{Easy} region is defined as:
\[
\mathcal{E}
=
\left\{
x :
\mathbb{E}_{y\sim\pi_\theta(\cdot|x)}
[r(x,y)]
\geq \tau
\right\},
\]

while the \textbf{Hard} region is defined as:
\[
\mathcal{H}
=
\left\{
x :
\mathbb{E}_{y\sim\pi_\theta(\cdot|x)}
[r(x,y)]
< \tau
\right\}.
\]

The on-policy GRPO gradient can then be decomposed as:
\begin{equation}
\begin{aligned}
\nabla_\theta J_{\text{on}}(\theta)
&=
p(\mathcal{E})\,\mathbb{E}_{\mathcal{E}}[g(x,y)] \\
&\quad +
p(\mathcal{H})\,\mathbb{E}_{\mathcal{H}}[g(x,y)].
\end{aligned}
\label{eq:on_policy_split}
\end{equation}
For Easy cases, the model already performs well during the early stage of training. The rewards within each group are generally high and close to each other, resulting in \(\hat A \approx 0\). Consequently, during the later stage of training, the optimization process mainly relies on the within-group relative advantages of Hard cases.

However, continuing to perform sampling on Easy cases at this stage becomes inefficient and leads to unnecessary computational overhead. Motivated by this observation, we propose \method, an improved variant built upon RULE. 

\subsection{The Pipeline of \method}
\begin{figure*}[t]
    \vspace{-0.3cm}
    \centering
    \includegraphics[width=\textwidth]{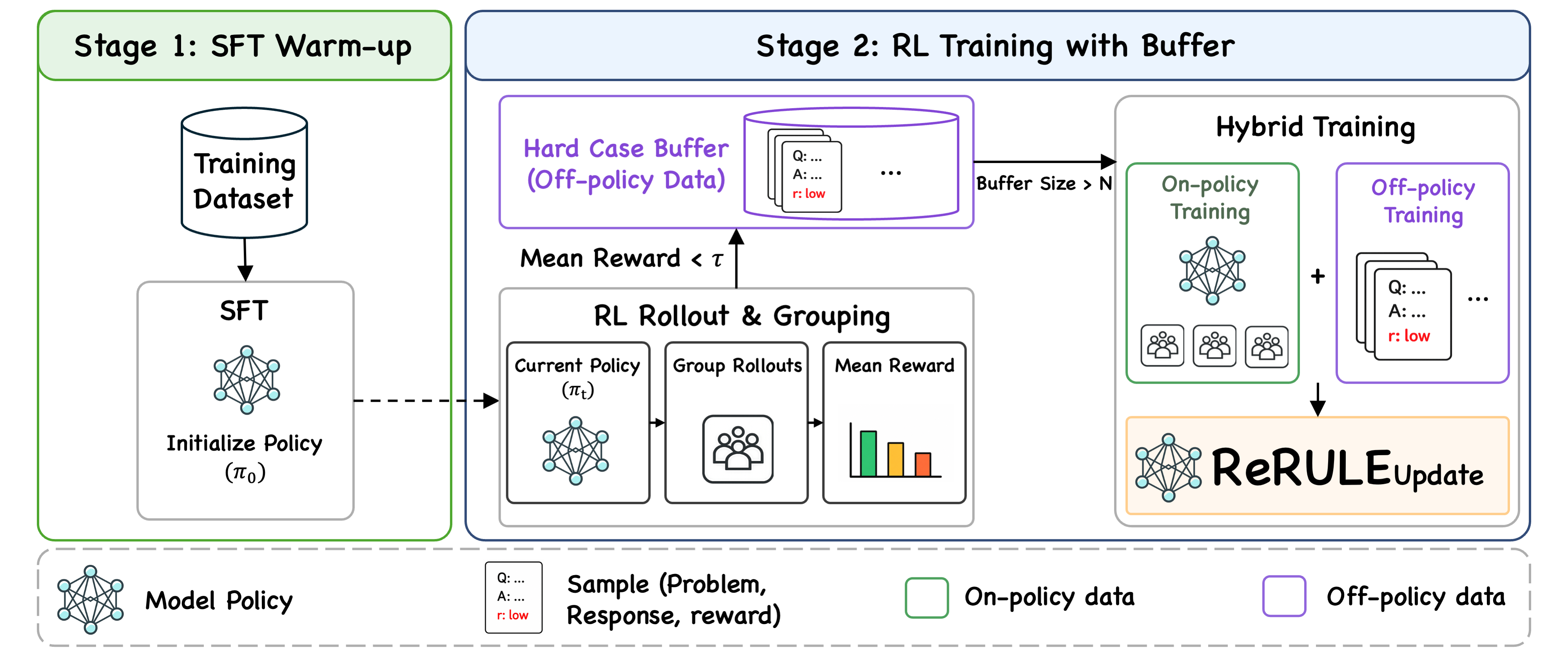}
    \caption{
Workflow of \method. Stage 1 follows RULE and performs SFT on a refusal dataset to initialize the model. In Stage 2, a hard case buffer is introduced: groups with mean reward below a threshold $\tau$ are stored during early GRPO training. When a certain condition is met, hybrid training is activated by combining on-policy rollouts with off-policy data sampled from the buffer.
}
    \label{fig:pipeline}

\vspace{-0.3cm}
\end{figure*}

\method mainly introduces modifications to Stage 2 of the original RULE framework. The Stage 1 SFT procedure remains identical to the original RULE framework, while the following modifications are introduced in Stage 2:

\begin{algorithm}[t]
\footnotesize
\caption{Stage 2 of \method}
\label{alg:hybrid}
\begin{algorithmic}[1]

\Require Policy $\pi_\theta$, replay buffer $\mathcal{B}$, threshold $\tau$, 
minimum buffer size $B_{\min}$, warm-up step $T_{\mathrm{warm}}$

\For{each training step $t$}

    \State Sample prompts from $\mathcal{D}_f \cup \widetilde{\mathcal{D}}_r$
    \State Generate rollout groups $g$ with policy $\pi_\theta$
    \State Compute rewards $r(x,y)$ and GRPO advantages $\hat A(x,y)$
    \State Perform on-policy GRPO update using $\nabla_\theta J_{\text{on}}(\theta)$

    \For{each rollout group $g$}
        \If{$\bar r(g) < \tau$}
            \State Store $g$ into replay buffer $\mathcal{B}$
        \EndIf
    \EndFor

    \If{$|\mathcal{B}| \ge B_{\min}$ and $t \ge T_{\mathrm{warm}}$}
        \State Sample replay groups from $\mathcal{B}$
        \State Compute Importance Sampling(IS) weight $\tilde\rho(y|x)$
        \State Perform off-policy replay update using $\nabla_\theta J_{\text{off}}(\theta)$
    \EndIf

\EndFor

\end{algorithmic}

\end{algorithm}

During the early stage of training, we perform standard on-policy GRPO optimization while simultaneously constructing a replay buffer that stores groups whose mean rewards are lower than a predefined threshold \(\tau\). Intuitively, these groups correspond to cases where the model exhibits unsatisfactory sampling behavior during the early training stage.

\vspace{-0.2cm}
During the later stage of training, in addition to the standard on-policy GRPO update, we introduce an additional off-policy replay update at each training step based on samples drawn from the replay buffer. Specifically, the off-policy gradient is defined as:
\vspace{-0.2cm}
\begin{equation}
\vspace{-0.2cm}
\begin{aligned}
\nabla_\theta J_{\text{off}}(\theta)
&=
\mathbb{E}_{g\sim\mathcal{B}}
\Bigl[
\mathbb{E}_{y\sim\pi_{\theta_{\text{old}}}}
\Bigl[
\\
&\quad
\tilde\rho(y|x)
\nabla_\theta \log \pi_\theta(y|x)
\hat A(x,y)
\Bigr]
\Bigr]
\end{aligned}
\label{eq:off_policy_gradient}
\end{equation}
where \(\mathcal{B}\) denotes the replay buffer, and
{\footnotesize
\[
\tilde\rho(y|x)
=
\operatorname{Normalize}
\left(
\operatorname{Clip}
\left(
\frac{\pi_\theta(y|x)}
{\pi_{\theta_{\text{old}}}(y|x)}
\right)
\right)
\]
}
represents the importance-sampling weight used to reweight replayed trajectories under the current policy. The clipping range in \(\operatorname{Clip}(\cdot)\) is a tunable hyperparameter and can be set during the training configuration.

Such replay-based retraining on Hard cases stored in the replay buffer is expected to improve both the efficiency and performance of Stage 2 GRPO optimization in RULE. This motivation is empirically validated in later Section~\ref{exp_muse} and Figure~\ref{fig:musecompare}. We further provide a theoretical analysis showing that \method achieves a tighter convergence bound than pure on-policy RULE under the hard/easy case decomposition (Appendix~\ref{app:refusal_boundary}).

\section{Experiment}
In this section, we evaluate the performance of \method on tasks commonly studied in LLM unlearning: Real-World Knowledge Unlearning, Copyrighted Unlearning, and Entity Unlearning. We conduct experiments on the corresponding benchmarks: RWKU\citep{jin2024rwkubenchmarkingrealworldknowledge}, MUSE-Books\citep{shi2024muse}, and TOFU\citep{maini2024tofu}, respectively.

\subsection{Baseline Methods}
We compare \method with existing unlearning approaches to evaluate its effectiveness. Across all experiments, we include three representative methods, GA \citep{yao2024large}, NPO \citep{zhang2024negative}, and RULE \citep{zhang2025rulereinforcementunlearningachieves}. For the RWKU and MUSE-Books benchmarks, we additionally compare \method against Sim-NPO \citep{fan2025simplicityprevailsrethinkingnegative}, a variant of NPO. For the TOFU benchmark, we further include KL \citep{maini2024tofu}, GD \citep{liu2022continual}, LLMU \citep{yao2024large}, PO \citep{maini2024tofu}, DPO, FLAT \citep{wang2024llm}, and SGA \citep{pang2025labelsmoothingimprovesgradient} as baselines.

\subsection{Real-World Knowledge Unlearning}
\begin{table*}[!t]
\vspace{-0.2cm}
\centering
\resizebox{0.8\linewidth}{!}{
\begin{tabular}{l cc cccc cccc ccc}
\toprule
\multirow{2}{*}{\textbf{Methods}} &
\multicolumn{2}{c}{\textbf{\# Tokens}} &
\multicolumn{4}{c}{\textbf{Forget Quality (↓)}} &
\multicolumn{4}{c}{\textbf{Forget Naturalness (↑)}} &
\multicolumn{3}{c}{\textbf{Retain Quality (↑)}} \\
\cmidrule(lr){2-3} \cmidrule(lr){4-7} \cmidrule(lr){8-11} \cmidrule(lr){12-14}
 & $D_f$ & $D_r$ & FB & QA & AA & All & Read & Help & Truth & All & FB & QA & All \\
\midrule
\textbf{Original} & 0\% & 0\% & 85.6 & 70.3 & 74.7 & 76.9 & 94.0 & 26.4 & 91.5 & 70.6 & 93.1 & 82.0 & 87.6 \\
\midrule
\textbf{GA}   &     & 0\%   & 72.0 & 64.6 & 68.5 & 68.4 & 45.8 & 33.2 & 43.2 & 40.7 & \colorbox{cyan!15}{\textbf{85.0}} & \colorbox{cyan!15}{\textbf{74.7}} & \colorbox{cyan!15}{\textbf{79.8}} \\
\ \ +GDR      & 100\% & 100\% & 72.6 & 64.0 & 69.7 & 68.8 & 30.4 & 23.5 & 27.2 & 27.0 & \colorbox{green!15}{\textbf{86.2}} & \colorbox{green!15}{\textbf{76.5}} & \colorbox{green!15}{\textbf{81.4}} \\
\ \ +KLR      &   & 100\% & 70.7 & 57.5 & 69.9 & 66.1 & 39.7 & 27.6 & 33.1 & 33.5 & 80.5 & 70.5 & 75.5 \\
\midrule
\textbf{NPO}  &     & 0\%   & 46.6 & 39.0 & 35.3 & 40.3 & 39.9 & 25.9 & 36.3 & 34.0 & 79.2 & 70.9 & 75.1 \\
\ \ +GDR      & 100\% & 100\% & 52.2 & 43.9 & 42.9 & 46.3 & 89.7 & 56.2 & 67.7 & 71.2 & 82.5 & 70.5 & 76.5 \\
\ \ +KLR      &  & 100\% & 52.5 & 40.6 & 43.2 & 45.4 & 92.1 & 56.6 & 69.6 & 72.8 & 83.2 & 72.1 & 77.6 \\
\midrule
\textbf{SimNPO} &   & 0\%   & 42.1 & 36.1 & 42.2 & 40.1 & 35.5 & 26.4 & 29.6 & 30.5 & 82.8 & 70.3 & 76.5 \\
\ \ +GDR        & 100\% & 100\% & 51.1 & 39.2 & 50.7 & 47.0 & 39.4 & 23.9 & 29.7 & 31.0 & 83.6 & 75.3 & 79.5 \\
\ \ +KLR        &  & 100\% & 44.6 & 35.4 & 44.6 & 41.5 & 50.6 & 25.5 & 34.5 & 36.9 & 82.9 & 71.4 & 77.1 \\
\midrule

\multicolumn{14}{c}{\textbf{RULE}} \\
\hdashline
Rej.\ Steer   & 6.29\% & 0\%   & 77.1 & 43.0 & 51.2 & 57.1 & 90.7 & 34.8 & 94.8 & 73.4 & 83.2 & 71.6 & 77.4 \\
\textbf{RULE$_{\text{GRPO}}$} & 12.1\% & 8.03\% & \colorbox{cyan!15}{\textbf{28.0}} & \colorbox{cyan!15}{\textbf{16.8}} & \colorbox{cyan!15}{\textbf{38.3}} & \colorbox{cyan!15}{\textbf{27.7}} & \colorbox{green!15}{\textbf{99.6}} & \colorbox{green!15}{\textbf{71.9}} & \colorbox{green!15}{\textbf{95.7}} & \colorbox{green!15}{\textbf{89.1}} & 76.2 & 71.3 & 73.7 \\
\midrule
\rowcolor[RGB]{230,230,250}
\multicolumn{14}{c}{\textbf{\method}} \\
\hdashline
\textbf{\method$_{\text{GRPO}}$} & 12.1\% & 8.03\% & \colorbox{green!15}{\textbf{\underline{26.5}}}   & \colorbox{green!15}{\textbf{\underline{13.1}}}   & \colorbox{green!15}{\textbf{\underline{33.2}}}   & \colorbox{green!15}{\textbf{\underline{24.2}}}   & \colorbox{cyan!15}{\textbf{95.2}} & \colorbox{cyan!15}{\textbf{56.3}} & \colorbox{cyan!15}{\textbf{94.7}} & \colorbox{cyan!15}{\textbf{82.1}}  & \underline{79.5}  & 68.8   & \underline{74.1}   \\

\bottomrule

\end{tabular}
}
\caption{\textit{llama3-8b-instruct} results on RWKU. We also report the training tokens budget for $D_f$ and $D_r$ (Appendix~\ref{app:evaluation_metrics}). The best results are in \colorbox{green!15}{\textbf{green}} and the second best results are in \colorbox{cyan!15}{\textbf{blue}}. If \method achieves better results than RULE, \underline{underline it}.}
\vspace{-0.3cm}
\label{tab:rwku}
\end{table*}

\paragraph{Experiment Setup.}The RWKU benchmark \citep{jin2024rwkubenchmarkingrealworldknowledge} is a real-world knowledge unlearning benchmark consisting of 200 real-world celebrity entities as forgetting targets, with 13,131 multi-level forgetting probes in total. Unlike TOFU \citep{maini2024tofu}, RWKU follows a zero-shot unlearning setting, where neither the forget corpus nor the retain corpus is accessible during training. In our experiments, we follow the setting of RULE \citep{zhang2025rulereinforcementunlearningachieves} and use Meta Llama3-8B-Instruct \citep{grattafiori2024llama3herdmodels} as the base model. Boundary data are constructed by replacing target entities with neighboring entities, and only a small portion of the forget set and boundary set is used for reinforcement unlearning training.

\paragraph{Evaluation Metrics.}
To evaluate unlearning performance on RWKU, we follow RULE and report Forget Quality, Forget Naturalness, and Retain Quality. Forget Quality is measured by ROUGE-L on forget probes, including fill-in-the-blank (FB), question-answering (QA), and adversarial attack (AA) queries, where lower scores indicate better forgetting. Retain Quality is evaluated on neighboring retain probes, where higher scores indicate better preservation of non-target knowledge. In addition, Forget Naturalness measures the quality of refusal responses from the aspects of Readability, Helpfulness, and Truthfulness.

\paragraph{Under the same experimental settings (Appendix~\ref{app:exp_settings}), \method achieves better forgetting and retention performance than RULE.}
As shown in Table\ref{tab:rwku},under the GRPO training setting, \method achieves the best Forget Quality, surpassing RULE, while also maintaining Top-2 performance on Forget Naturalness. Moreover, \method outperforms RULE on Retain Quality, which is consistent with our argument in Section~\ref{method} that \method exhibits higher learning efficiency on hard cases. 

\subsection{Copyrighted Unlearning}
\label{exp_muse}
\paragraph{Experiment Setup.}
The MUSE benchmark \citep{shi2024muse} is a unlearning benchmark that requires models to forget copyrighted content while preserving general utility. Following RULE, we use Meta Llama2-7B as the base model. In the MUSE-books setting, the forgetting target is the Harry Potter corpus, which contains 3045 text passages. RULE constructs QA-style forget queries using GPT-4o-mini \citep{openai2024gpt4technicalreport} and synthesizes boundary data by replacing sensitive targets for subsequent reinforcement unlearning training.
\paragraph{Evaluation Metrics.}
To evaluate unlearning performance on MUSE-Books, we follow RULE and report three groups of metrics: Forget Quality, Forget Naturalness, and Retain Quality. Forget Quality includes VerbMem and KnowMem, which measure whether the model still memorizes the forgotten Harry Potter content, where lower scores indicate better forgetting. Forget Naturalness is evaluated from Readability, Helpfulness, and Truthfulness, with higher scores indicating more natural and reliable refusal responses. Retain Quality is measured by Utility, where higher values indicate better preservation of general model capability.

\begin{table*}[!t]

\centering
\small
\resizebox{0.8\textwidth}{!}{
\setlength{\tabcolsep}{6pt}
\renewcommand{\arraystretch}{1.15}
\begin{tabular}{lcccccccc}
\toprule
\multirow{2}{*}{\textbf{Methods}} 
& \multicolumn{2}{c}{\textbf{\# Tokens}} 
& \multicolumn{2}{c}{\textbf{Forget Quality($\downarrow$)}} 
& \multicolumn{3}{c}{\textbf{Forget Naturalness($\uparrow$)}} 
& \multicolumn{1}{c}{\textbf{Retain Quality($\uparrow$)}} \\
\cmidrule(lr){2-3}
\cmidrule(lr){4-5}
\cmidrule(lr){6-8}
\cmidrule(lr){9-9}
& $D_f$ & $D_r$ & Verb. & Know. & Read & Help & Truth & Utility \\
\midrule

\textbf{Original} 
& 0\% & 0\% & 58.4 & 63.9 & - & - & - & 55.2 \\
\midrule

\textbf{GA} 
& 100\% & 0\% & \colorbox{green!15}{\textbf{0.0}} & \colorbox{green!15}{\textbf{0.0}} & 94.0 & \colorbox{green!15}{\textbf{63.0}} & 77.6 & 0.0 \\
\quad +GDR 
& 100\% & 100\% & \colorbox{green!15}{\textbf{0.0}} & \colorbox{green!15}{\textbf{0.0}} & 94.0 & 60.0 & 79.6 & 10.9 \\
\quad +KLR 
& 100\% & 100\% & \colorbox{green!15}{\textbf{0.0}} & \colorbox{green!15}{\textbf{0.0}} & 94.0 & \colorbox{cyan!15}{\textbf{61.6}} & 80.0 & 40.5 \\
\midrule

\textbf{NPO} 
& 100\% & 0\% & 11.9 & 4.7 & 94.4 & 58.6 & 80.0 & 5.9 \\
\quad +GDR 
& 100\% & 100\% & 21.1 & 32.5 & 94.0 & 58.2 & 78.0 & 62.4 \\
\quad +KLR 
& 100\% & 100\% & 8.0 & 45.4 & 94.6 & 60.4 & 81.4 & \colorbox{green!15}{\textbf{67.3}} \\
\midrule

\textbf{SimNPO} 
& 100\% & 0\% & \colorbox{green!15}{\textbf{0.0}} & \colorbox{green!15}{\textbf{0.0}} & 93.8 & 60.2 & 80.6 & 0.0 \\
\quad +GDR 
& 100\% & 100\% & 0.6 & 23.4 & 95.2 & 59.6 & 81.2 & \colorbox{cyan!15}{\textbf{64.8}} \\
\quad +KLR 
& 100\% & 100\% & 47.4 & 46.2 & 94.6 & 61.2 & 82.4 & \colorbox{green!15}{\textbf{67.3}} \\
\midrule
\multicolumn{9}{c}{\textbf{RULE}} \\
\hdashline
\textbf{RULE$_{\text{GRPO}}$} 
& 2.9\% & 2.9\% & \colorbox{green!15}{\textbf{0.0}} & \colorbox{cyan!15}{\textbf{1.4}} &  \colorbox{green!15}{\textbf{99.2}}   &  38.2  &  \colorbox{cyan!15}{\textbf{90.8}} & 46.3 \\
\midrule
\rowcolor[RGB]{230,230,250}
\multicolumn{9}{c}{\textbf{\method}} \\
\hdashline
\textbf{\method$_{\text{GRPO}}$} 
& 2.9\% & 2.9\% & \colorbox{green!15}{\textbf{0.0}} &\colorbox{cyan!15}{\textbf{1.4}} & \colorbox{cyan!15}{\textbf{98.8}} & 28.7 & \colorbox{green!15}{\textbf{\underline{93.0}}} & \underline{56.2} \\

\bottomrule
\end{tabular}
}
\caption{\textbf{llama2-7b results on MUSE-books.} We report forgetting quality, naturalness of refusal, and utility retention. The training token ratio for $D_f$ and $D_r$ is listed per method (Appendix~\ref{app:evaluation_metrics}), the best results are in \colorbox{green!15}{\textbf{green}} and the second best results are in \colorbox{cyan!15}{\textbf{blue}}.If \method achieves better results than RULE, \underline{underline it}.
}
\label{tab:muse_books_llama2_7b}
\vspace{-0.3cm}
\end{table*}

\paragraph{\method achieves top-2 performance on both Forget Quality and Forget Naturalness.}As shown in Table~\ref{tab:muse_books_llama2_7b}, both RULE and \method achieve top-2 performance on most metrics. Moreover, compared with several methods that obtain the best Forget Quality, \method achieves a substantially higher Retain Quality score of 56.2, while the original RULE only achieves 46.3.

\begin{figure}[t]
    \centering

    \begin{subfigure}{1\linewidth}
        \centering
        \includegraphics[width=1\linewidth]{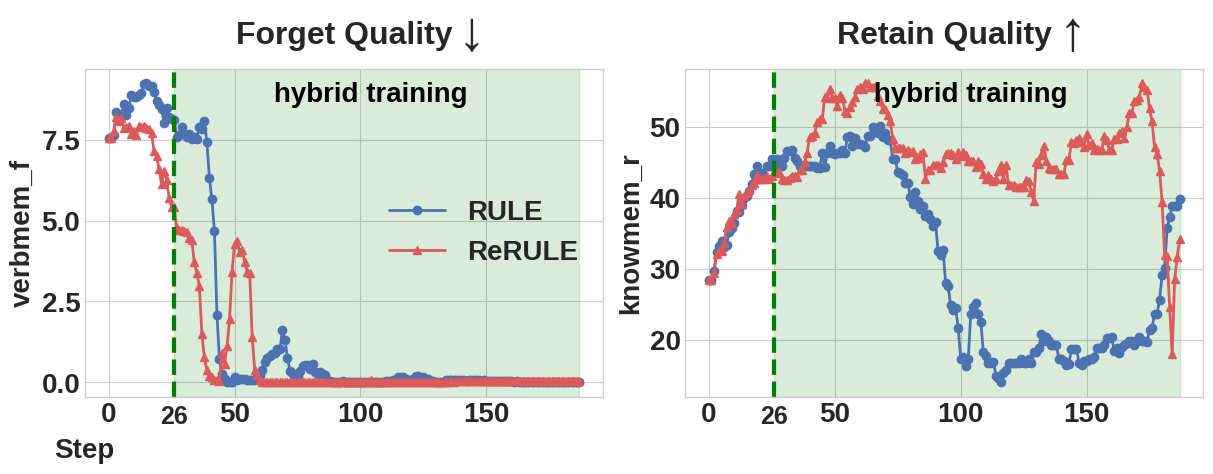}
        \caption{Overall metric comparison on MUSE.}
        \label{fig:musecompare_a}
    \end{subfigure}

    \vspace{0.5em}

    \begin{subfigure}{1\linewidth}
        \centering
        \includegraphics[width=1\linewidth]{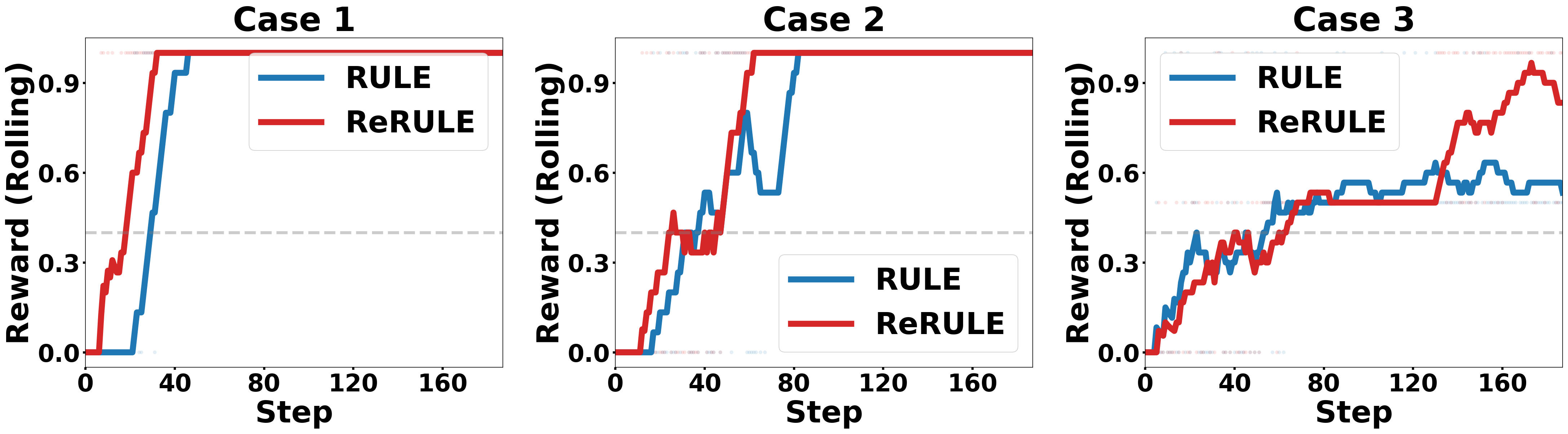}
        \caption{Tracking of hard cases during training.}
        \label{fig:musecompare_b}
    \end{subfigure}

    \caption{(a) Comparison between RULE and \method on MUSE. The green vertical line marks the start of hybrid training at step 26. (b) Smoothed reward curves for three hard cases whose early-stage average reward is below 0.4.}
    \label{fig:musecompare}
    \vspace{-0.5cm}
\end{figure}

\paragraph{Improved Retain Quality demonstrates the effectiveness of \method on hard cases.}
As shown in Figure~\ref{fig:musecompare_a}, both RULE and \method quickly achieve strong Forget Quality, partly because forget-related tasks in MUSE-Books are relatively easy, as shown in Figure~\ref{fig:sampling_a}. 
However, RULE shows substantial instability in Retain Quality, suggesting that it keeps exploring hard cases but rarely samples responses that effectively improve KnowMem\_r. 
In contrast, after hybrid off-policy/on-policy training starts at step 26, \method increases learning on hard cases and consistently achieves better Retain Quality under the same number of training steps, consistent with our analysis in Section~\ref{method}.

\paragraph{On specific hard cases, \method outperforms RULE.}
Table~\ref{tab:hard_cases} lists three hard cases from MUSE-Books, with reward curves shown in Figure~\ref{fig:musecompare_b}. 
Under \method training, these cases often achieve higher rewards than under RULE.
\begin{table}[t]
\vspace{-0.3cm}
\centering
\footnotesize
\setlength{\tabcolsep}{3pt}
\renewcommand{\arraystretch}{1.15}
\begin{tabularx}{\linewidth}{c>{\hsize=1.65\hsize}X>{\hsize=0.35\hsize}Xc}
\toprule
Case & Question & Answer & Type \\
\midrule
1
& What phrase was Harry instructed to write repeatedly with the special quill given by Umbridge?
& I must not tell lies
& reject \\
\midrule
2
& How long has Snape been teaching at Hogwarts?
& Fourteen years
& reject \\
\midrule
3
& What was the name of the dragon that Hagrid illegally kept and sought help from Charlie Weasley to get rid of?
& Norbert
& normal \\
\bottomrule
\end{tabularx}
\caption{Examples of hard cases with their corresponding answers and response types.}
\label{tab:hard_cases}
\vspace{-0.4cm}
\end{table}

\paragraph{When the reply buffer stores random samples instead of hard cases, \method degenerates into RULE.}
To examine the role of hard cases, we conduct a controlled experiment where the reply buffer randomly stores samples during training, followed by the same hybrid training in the later stage. The results are reported as $\method_{\text{random}}$ in Table~\ref{tab:random_rb}. When the storage strategy becomes random, $\method_{\text{random}}$ degrades to a level close to RULE. This matches the motivation of \method: random storage repeatedly trains the model on selected samples without considering its performance on them.
\begin{table}[h]
\centering
\vspace{-0.2cm}
\resizebox{0.75\linewidth}{!}{
\begin{tabular}{lccc}
\toprule
Methods
& \multicolumn{2}{c}{Forget Quality ($\downarrow$)}
& Retain Quality ($\uparrow$) \\
\cmidrule(lr){2-3}
\cmidrule(lr){4-4}
& Verb. & Know. & Utility \\
\midrule

RULE
& 0.0
& 1.4
& 46.3 \\

\method
& 0.0
& 1.4
& 56.2 \\

$\method_{\text{random}}$
& 0.0
& 1.5
& 46.3 \\

\bottomrule
\end{tabular}
}
\caption{A controlled experiment with random storage in the reply buffer.}
\label{tab:random_rb}
\vspace{-0.3cm}
\end{table}

\subsection{Entity Unlearning}
\paragraph{Experiment Setup.}
The TOFU benchmark \citep{maini2024tofu} is a question-answering dataset built on the biographies of 200 fictional authors, with each author associated with 20 QA pairs. Based on the designated unlearning scope, the dataset is divided into a forget set and a retain set. In our experiments, we adopt the 1\% forget split. Following prior work \citep{pang2025labelsmoothingimprovesgradient}, we use Meta Llama2-7B \citep{touvron2023llama} as the base LLM for all experiments.

\paragraph{Evaluation Metrics.}
To evaluate the forgetting performance and retained utility of unlearned models on TOFU, we report Forget Quality (FQ), Model Utility (MU), and Forget ROUGE-L (F-RL) following prior work \citep{maini2024tofu}. FQ is computed using the Kolmogorov–Smirnov test between the unlearned model and the retain-only model, where higher values indicate better forgetting. MU measures the retained utility on held-out retain data, where higher values are preferred. We additionally report F-RL on the forget set, where lower scores indicate better forgetting performance.

\begin{table}[h]
\vspace{-0.2cm}
\centering
\scriptsize  

\begin{tabular}{l ccc}
\toprule
\multirow{2}{*}{\textbf{Model}} &
\multicolumn{3}{c}{\textbf{Llama2-7B}} \\
\cmidrule(lr){2-4}
& FQ(\(\uparrow\)) & MU(\(\uparrow\)) & F-RL(\(\downarrow\))  \\
\midrule

\textbf{Original LLM} & 4.4883e-06 & 0.6239 & 0.9851  \\
\textbf{Retained LLM} & 1.0000 & 0.6267 & 0.4080  \\
\midrule

GA/SGA (r = 0) & \colorbox{cyan!15}{\textbf{0.0068}} & 0.5990 & 0.4817  \\
KL             & 0.0030 & 0.5994 & 0.4922  \\
GD             & \colorbox{cyan!15}{\textbf{0.0068}} & 0.5998 & 0.4869  \\
LLMU           & 0.0030 & 0.5999 & 0.4891  \\
PO             & 0.0030 & \colorbox{green!15}{\textbf{0.6323}} & \colorbox{cyan!15}{\textbf{0.1752}} \\
DPO-RT         & \colorbox{cyan!15}{\textbf{0.0068}} & 0.6322 & 0.2595 \\
NPO-RT         & 0.0030 & 0.5994 & 0.5049 \\
FLAT (Pearson) & 0.0030 & \colorbox{cyan!15}{\textbf{0.6304}} & 0.4825 \\
SGA (r = 0.4)  & \colorbox{cyan!15}{\textbf{0.0068}} & 0.6027 & 0.4792  \\
\midrule
\multicolumn{4}{c}{\textbf{RULE}} \\
\hdashline
\textbf{RULE$_{\text{GRPO}}$} 
& \colorbox{green!15}{\textbf{0.1650}} &0.6151 & \colorbox{green!15}{\textbf{0.1322}}  \\
\midrule
\rowcolor[RGB]{230,230,250}
\multicolumn{4}{c}{\textbf{\method}} \\
\hdashline
\textbf{\method$_{\text{GRPO}}$} 
& \colorbox{green!15}{\textbf{0.1650}} & 0.6076 & \colorbox{green!15}{\textbf{0.1322}} \\
\bottomrule
\end{tabular}

\caption{Results of \textit{Llama2-7B} on TOFU. FQ, MU, F-RL, and R-RL denote Forget Quality, Model Utility, and ROUGE-L scores on the forget and retain sets, respectively. The best results are in \colorbox{green!15}{\textbf{green}} and the second best results are in \colorbox{cyan!15}{\textbf{blue}}}
\label{tab:tofu-llama2}
\vspace{-0.35cm}
\end{table}

\paragraph{Both \method and RULE achieve strong performance.
}Compared with other baselines, both \method and RULE achieve the best Forget Quality and Forget ROUGE-L while maintaining strong Model Utility, showing a favorable FQ-MU trade-off~\citep{pang2025labelsmoothingimprovesgradient}. This is because they reformulate unlearning as learning precise refusal behaviors rather than directly erasing knowledge. However, \method brings limited improvement over RULE on TOFU, since the TOFU forget01 setting is relatively small and easy. As shown in Table~\ref{tab:tau_ratio}, TOFU contains fewer hard cases during early training, leaving limited room for replay-based learning.

\begin{table}[h]
\centering
\footnotesize
\vspace{-0.2cm}
\resizebox{0.55\linewidth}{!}{
\begin{tabular}{lcc}
\toprule
Experiment & $\tau$ & Hard Ratio \\
\midrule
MUSE & 0.4 & 67.5 \\
TOFU & 0.4 & 47.5 \\
TOFU & 0.6 & 54.3 \\
\bottomrule
\end{tabular}
}
\caption{Hard ratios under different threshold settings.}
\label{tab:tau_ratio}
\vspace{-0.3cm}
\end{table}

\subsection{Ablation Study}
\paragraph{Regarding computational cost.}
Since \method adds an extra replay-based gradient update with importance ratios after each standard on-policy step, its computational cost may raise concerns. However, the replay buffer stores both hard-case prompts and their sampled rollouts, avoiding repeated sampling during later off-policy training. Thus, \method requires only modest additional computation. We report the training time of different experiments in Table~\ref{tab:cost}.

\begin{table}[h]
\centering
\footnotesize
\vspace{-0.2cm}
\resizebox{0.8\linewidth}{!}{
\begin{tabular}{lccc}
\toprule
\textbf{Benchmarks} & \textbf{RWKU} & \textbf{MUSE} & \textbf{TOFU} \\
\midrule
Data Size & 12000 & 5988 & 80 \\
\midrule
\multicolumn{4}{c}{\textbf{Training Time}} \\
RULE & 21.2 h & 6.8 h & 1.8 h \\
\method & 22.3 h & 7.1 h & 2.0 h \\
\bottomrule
\end{tabular}
}
\caption{Training cost comparison between RULE and \method on different benchmarks. All training experiments were conducted on 4 NVIDIA A800 GPUs.}
\label{tab:cost}
\vspace{-0.3cm}
\end{table}

\paragraph{Regarding the setting of the reply buffer.}To examine whether the reply buffer design is a key factor affecting the experimental results, we conduct ablation studies with different values of $\tau$, which controls the criterion for selecting data into the reply buffer. Because of the design of RULE’s reward function, the reward of each sampled output can only take one of three values: 0, 0.5, or 1.0. Therefore, when $\tau$ is set to 0.4, only cases with reward 0 are stored in the replay buffer; when $\tau$ is set to 0.6, cases with rewards 0 and 0.5 are stored, and so on. The results are shown in Table~\ref{tab:muse_books_llama2_7b_simple}.

\begin{table}[H]
\footnotesize
\centering
\vspace{-0.2cm}
\resizebox{0.85\linewidth}{!}{
\begin{tabular}{lccc}
\toprule
Methods
& \multicolumn{2}{c}{Forget Quality ($\downarrow$)}
& Retain Quality ($\uparrow$) \\
\cmidrule(lr){2-3}
\cmidrule(lr){4-4}
& Verb. & Know. & Utility \\
\midrule

RULE
& 0.0
& 1.4
& 46.3 \\

\method
& 0.0
& 1.4
& 56.2 \\

$\method_{0.6}$
& 6.6
& 1.3
& 52.3 \\

$\method_{1.0}$
& 4.3
& 1.4
& 53.7 \\

\bottomrule
\end{tabular}
}
\caption{Experimental results under different reply buffer settings. $\method_{0.6}$ denotes the variant where the threshold $\tau$ is set to 0.6, while $\method_{1.0}$ denotes the variant where $\tau$ is set to 1.0. In the original MUSE-Books experiment of \method, $\tau$ is set to 0.4.}
\label{tab:muse_books_llama2_7b_simple}
\vspace{-0.2cm}
\end{table}
\vspace{-0.2cm}
As $\tau$ increases, the criterion for hard cases becomes less selective, causing more samples that are already relatively easy for the model to be stored in the reply buffer. During hybrid training, repeatedly updating on these easy samples can introduce redundant optimization, reducing training efficiency and ultimately leading to worse overall performance.

\section{Conclusion}
Building on RULE, this paper proposes \method, an off-policy RL enhancement for LLM unlearning. \method stores hard cases encountered during GRPO training in a replay buffer and later combines on-policy GRPO with importance-weighted off-policy updates, improving learning efficiency on hard cases. Experiments on multiple benchmarks show that \method achieves competitive performance among fine-tuning-based unlearning methods, and can outperform RULE in hard-case scenarios with only modest additional computation.

\section*{Limitations}
This paper mainly targets a specific issue in the unlearning setting: within the RULE framework, the Reject and Normal prompts often exhibit significantly different levels of difficulty, which leads to relatively low training efficiency during the later stage of RULE training.

For datasets where prompt difficulty is relatively uniform or generally easy, the improvement brought by \method is expected to be very limited. This phenomenon can also be observed from the results on the TOFU task presented in our paper.

\section*{Ethical Considerations}
Machine unlearning is closely connected to ethical issues such as privacy protection, the right to erasure, copyright protection, and responsible model deployment. Although the experiments in this paper do not involve human subjects or newly collected sensitive personal data, we acknowledge that unlearning methods may have broader ethical implications. First, incomplete unlearning may create a false sense of privacy or copyright compliance if residual memorization, paraphrased knowledge, or membership signals remain in the model. Second, overly aggressive unlearning may cause collateral degradation of useful retained knowledge, especially when forget and retain data are semantically entangled. Third, unlearning techniques are dual-use: while they can help remove private, copyrighted, biased, or unsafe information, they may also be misused to suppress information or obscure accountability. 

In our experiments, TOFU uses fictional personal profiles constructed by its authors, RWKU is based on well-known real-world facts, and MUSE-Books focuses on copyrighted book-related information. We use these benchmarks only for research evaluation and do not attempt to reconstruct, redistribute, or expose private or copyrighted content. Our results should therefore be interpreted as empirical evidence about unlearning behavior under benchmark settings, rather than as a guarantee of complete legal or privacy compliance in real-world deployment.

\clearpage
\bibliography{custom}

@misc{zhang2025rulereinforcementunlearningachieves,
      title={RULE: Reinforcement UnLEarning Achieves Forget-Retain Pareto Optimality},
      author={Chenlong Zhang and Zhuoran Jin and Hongbang Yuan and Jiaheng Wei and Tong Zhou and Kang Liu and Jun Zhao and Yubo Chen},
      year={2025},
      eprint={2506.07171},
      archivePrefix={arXiv},
      primaryClass={cs.CL},
      url={https://arxiv.org/abs/2506.07171}
}

@misc{shao2024deepseekmathpushinglimitsmathematical,
      title={DeepSeekMath: Pushing the Limits of Mathematical Reasoning in Open Language Models}, 
      author={Zhihong Shao and Peiyi Wang and Qihao Zhu and Runxin Xu and Junxiao Song and Xiao Bi and Haowei Zhang and Mingchuan Zhang and Y. K. Li and Y. Wu and Daya Guo},
      year={2024},
      eprint={2402.03300},
      archivePrefix={arXiv},
      primaryClass={cs.CL},
      url={https://arxiv.org/abs/2402.03300}, 
}

@misc{schulman2017proximalpolicyoptimizationalgorithms,
      title={Proximal Policy Optimization Algorithms}, 
      author={John Schulman and Filip Wolski and Prafulla Dhariwal and Alec Radford and Oleg Klimov},
      year={2017},
      eprint={1707.06347},
      archivePrefix={arXiv},
      primaryClass={cs.LG},
      url={https://arxiv.org/abs/1707.06347}, 
}

@article{maini2024tofu,
  title={Tofu: A task of fictitious unlearning for llms},
  author={Maini, Pratyush and Feng, Zhili and Schwarzschild, Avi and Lipton, Zachary C and Kolter, J Zico},
  journal={arXiv preprint arXiv:2401.06121},
  year={2024}
}

@misc{jin2024rwkubenchmarkingrealworldknowledge,
      title={RWKU: Benchmarking Real-World Knowledge Unlearning for Large Language Models}, 
      author={Zhuoran Jin and Pengfei Cao and Chenhao Wang and Zhitao He and Hongbang Yuan and Jiachun Li and Yubo Chen and Kang Liu and Jun Zhao},
      year={2024},
      eprint={2406.10890},
      archivePrefix={arXiv},
      primaryClass={cs.CL},
      url={https://arxiv.org/abs/2406.10890}, 
}

@article{shi2024muse,
  title={Muse: Machine unlearning six-way evaluation for language models},
  author={Shi, Weijia and Lee, Jaechan and Huang, Yangsibo and Malladi, Sadhika and Zhao, Jieyu and Holtzman, Ari and Liu, Daogao and Zettlemoyer, Luke and Smith, Noah A and Zhang, Chiyuan},
  journal={arXiv preprint arXiv:2407.06460},
  year={2024}
}

@inproceedings{liu2022continual,
  title={Continual learning and private unlearning},
  author={Liu, Bo and Liu, Qiang and Stone, Peter},
  booktitle={Conference on Lifelong Learning Agents},
  pages={243--254},
  year={2022},
  organization={PMLR}
}

@article{wang2024llm,
  title={Llm unlearning via loss adjustment with only forget data},
  author={Wang, Yaxuan and Wei, Jiaheng and Liu, Chris Yuhao and Pang, Jinlong and Liu, Quan and Shah, Ankit Parag and Bao, Yujia and Liu, Yang and Wei, Wei},
  journal={arXiv preprint arXiv:2410.11143},
  year={2024}
}

@article{zhang2024negative,
  title={Negative preference optimization: From catastrophic collapse to effective unlearning},
  author={Zhang, Ruiqi and Lin, Licong and Bai, Yu and Mei, Song},
  journal={arXiv preprint arXiv:2404.05868},
  year={2024}
}

@misc{fan2025simplicityprevailsrethinkingnegative,
      title={Simplicity Prevails: Rethinking Negative Preference Optimization for LLM Unlearning}, 
      author={Chongyu Fan and Jiancheng Liu and Licong Lin and Jinghan Jia and Ruiqi Zhang and Song Mei and Sijia Liu},
      year={2025},
      eprint={2410.07163},
      archivePrefix={arXiv},
      primaryClass={cs.CL},
      url={https://arxiv.org/abs/2410.07163}, 
}

@article{yao2024large,
  title={Large language model unlearning},
  author={Yao, Yuanshun and Xu, Xiaojun},
  journal={Advances in Neural Information Processing Systems},
  volume={37},
  pages={105425--105475},
  year={2024}
}

@misc{pang2025labelsmoothingimprovesgradient,
      title={Label Smoothing Improves Gradient Ascent in LLM Unlearning}, 
      author={Zirui Pang and Hao Zheng and Zhijie Deng and Ling Li and Zixin Zhong and Jiaheng Wei},
      year={2025},
      eprint={2510.22376},
      archivePrefix={arXiv},
      primaryClass={cs.LG},
      url={https://arxiv.org/abs/2510.22376}, 
}

@misc{yang2025qwen3technicalreport,
      title={Qwen3 Technical Report}, 
      author={An Yang and Anfeng Li and Baosong Yang and Beichen Zhang and Binyuan Hui and Bo Zheng and Bowen Yu and Chang Gao and Chengen Huang and Chenxu Lv and Chujie Zheng and Dayiheng Liu and Fan Zhou and Fei Huang and Feng Hu and Hao Ge and Haoran Wei and Huan Lin and Jialong Tang and Jian Yang and Jianhong Tu and Jianwei Zhang and Jianxin Yang and Jiaxi Yang and Jing Zhou and Jingren Zhou and Junyang Lin and Kai Dang and Keqin Bao and Kexin Yang and Le Yu and Lianghao Deng and Mei Li and Mingfeng Xue and Mingze Li and Pei Zhang and Peng Wang and Qin Zhu and Rui Men and Ruize Gao and Shixuan Liu and Shuang Luo and Tianhao Li and Tianyi Tang and Wenbiao Yin and Xingzhang Ren and Xinyu Wang and Xinyu Zhang and Xuancheng Ren and Yang Fan and Yang Su and Yichang Zhang and Yinger Zhang and Yu Wan and Yuqiong Liu and Zekun Wang and Zeyu Cui and Zhenru Zhang and Zhipeng Zhou and Zihan Qiu},
      year={2025},
      eprint={2505.09388},
      archivePrefix={arXiv},
      primaryClass={cs.CL},
      url={https://arxiv.org/abs/2505.09388}, 
}

@misc{deng2025guardgenerationtimellmunlearning,
      title={GUARD: Generation-time LLM Unlearning via Adaptive Restriction and Detection}, 
      author={Zhijie Deng and Chris Yuhao Liu and Zirui Pang and Xinlei He and Lei Feng and Qi Xuan and Zhaowei Zhu and Jiaheng Wei},
      year={2025},
      eprint={2505.13312},
      archivePrefix={arXiv},
      primaryClass={cs.CL},
      url={https://arxiv.org/abs/2505.13312}, 
}

@article{pawelczyk2023context,
  title={In-context unlearning: Language models as few shot unlearners},
  author={Pawelczyk, Martin and Neel, Seth and Lakkaraju, Himabindu},
  journal={arXiv preprint arXiv:2310.07579},
  year={2023}
}

@article{muresanu2024fast,
  title={Fast exact unlearning for in-context learning data for LLMs},
  author={Muresanu, Andrei I and Thudi, Anvith and Zhang, Michael R and Papernot, Nicolas},
  journal={arXiv preprint arXiv:2402.00751},
  year={2024}
}

@article{thaker2024guardrail,
  title={Guardrail baselines for unlearning in llms},
  author={Thaker, Pratiksha and Maurya, Yash and Hu, Shengyuan and Wu, Zhiwei Steven and Smith, Virginia},
  journal={arXiv preprint arXiv:2403.03329},
  year={2024}
}

@article{gao2024large,
  title={On large language model continual unlearning},
  author={Gao, Chongyang and Wang, Lixu and Ding, Kaize and Weng, Chenkai and Wang, Xiao and Zhu, Qi},
  journal={arXiv preprint arXiv:2407.10223},
  year={2024}
}

@article{liu2024large,
  title={Large language model unlearning via embedding-corrupted prompts},
  author={Liu, Chris Y and Wang, Yaxuan and Flanigan, Jeffrey and Liu, Yang},
  journal={Advances in Neural Information Processing Systems},
  volume={37},
  pages={118198--118266},
  year={2024}
}

@inproceedings{bhaila2025soft,
  title={Soft prompting for unlearning in large language models},
  author={Bhaila, Karuna and Van, Minh-Hao and Wu, Xintao},
  booktitle={Proceedings of the 2025 Conference of the Nations of the Americas Chapter of the Association for Computational Linguistics: Human Language Technologies (Volume 1: Long Papers)},
  pages={4046--4056},
  year={2025}
}

@inproceedings{bourtoule2021machine,
  title={Machine unlearning},
  author={Bourtoule, Lucas and Chandrasekaran, Varun and Choquette-Choo, Christopher A and Jia, Hengrui and Travers, Adelin and Zhang, Baiwu and Lie, David and Papernot, Nicolas},
  booktitle={2021 IEEE symposium on security and privacy (SP)},
  pages={141--159},
  year={2021},
  organization={IEEE}
}

@inproceedings{wang2023kga,
  title={Kga: A general machine unlearning framework based on knowledge gap alignment},
  author={Wang, Lingzhi and Chen, Tong and Yuan, Wei and Zeng, Xingshan and Wong, Kam-Fai and Yin, Hongzhi},
  booktitle={Proceedings of the 61st Annual Meeting of the Association for Computational Linguistics (Volume 1: Long Papers)},
  pages={13264--13276},
  year={2023}
}

@article{ethayarajh2024kto,
  title={Kto: Model alignment as prospect theoretic optimization},
  author={Ethayarajh, Kawin and Xu, Winnie and Muennighoff, Niklas and Jurafsky, Dan and Kiela, Douwe},
  journal={arXiv preprint arXiv:2402.01306},
  year={2024}
}

@misc{grattafiori2024llama3herdmodels,
      title={The Llama 3 Herd of Models}, 
      author={Aaron Grattafiori and Abhimanyu Dubey and Abhinav Jauhri and Abhinav Pandey and Abhishek Kadian and Ahmad Al-Dahle and Aiesha Letman and Akhil Mathur and Alan Schelten and Alex Vaughan and Amy Yang and Angela Fan and Anirudh Goyal and Anthony Hartshorn and Aobo Yang and Archi Mitra and Archie Sravankumar and Artem Korenev and Arthur Hinsvark and Arun Rao and Aston Zhang and Aurelien Rodriguez and Austen Gregerson and Ava Spataru and Baptiste Roziere and Bethany Biron and Binh Tang and Bobbie Chern and Charlotte Caucheteux and Chaya Nayak and Chloe Bi and Chris Marra and Chris McConnell and Christian Keller and Christophe Touret and Chunyang Wu and Corinne Wong and Cristian Canton Ferrer and Cyrus Nikolaidis and Damien Allonsius and Daniel Song and Danielle Pintz and Danny Livshits and Danny Wyatt and David Esiobu and Dhruv Choudhary and Dhruv Mahajan and Diego Garcia-Olano and Diego Perino and Dieuwke Hupkes and Egor Lakomkin and Ehab AlBadawy and Elina Lobanova and Emily Dinan and Eric Michael Smith and Filip Radenovic and Francisco Guzmán and Frank Zhang and Gabriel Synnaeve and Gabrielle Lee and Georgia Lewis Anderson and Govind Thattai and Graeme Nail and Gregoire Mialon and Guan Pang and Guillem Cucurell and Hailey Nguyen and Hannah Korevaar and Hu Xu and Hugo Touvron and Iliyan Zarov and Imanol Arrieta Ibarra and Isabel Kloumann and Ishan Misra and Ivan Evtimov and Jack Zhang and Jade Copet and Jaewon Lee and Jan Geffert and Jana Vranes and Jason Park and Jay Mahadeokar and Jeet Shah and Jelmer van der Linde and Jennifer Billock and Jenny Hong and Jenya Lee and Jeremy Fu and Jianfeng Chi and Jianyu Huang and Jiawen Liu and Jie Wang and Jiecao Yu and Joanna Bitton and Joe Spisak and Jongsoo Park and Joseph Rocca and Joshua Johnstun and Joshua Saxe and Junteng Jia and Kalyan Vasuden Alwala and Karthik Prasad and Kartikeya Upasani and Kate Plawiak and Ke Li and Kenneth Heafield and Kevin Stone and Khalid El-Arini and Krithika Iyer and Kshitiz Malik and Kuenley Chiu and Kunal Bhalla and Kushal Lakhotia and Lauren Rantala-Yeary and Laurens van der Maaten and Lawrence Chen and Liang Tan and Liz Jenkins and Louis Martin and Lovish Madaan and Lubo Malo and Lukas Blecher and Lukas Landzaat and Luke de Oliveira and Madeline Muzzi and Mahesh Pasupuleti and Mannat Singh and Manohar Paluri and Marcin Kardas and Maria Tsimpoukelli and Mathew Oldham and Mathieu Rita and Maya Pavlova and Melanie Kambadur and Mike Lewis and Min Si and Mitesh Kumar Singh and Mona Hassan and Naman Goyal and Narjes Torabi and Nikolay Bashlykov and Nikolay Bogoychev and Niladri Chatterji and Ning Zhang and Olivier Duchenne and Onur Çelebi and Patrick Alrassy and Pengchuan Zhang and Pengwei Li and Petar Vasic and Peter Weng and Prajjwal Bhargava and Pratik Dubal and Praveen Krishnan and Punit Singh Koura and Puxin Xu and Qing He and Qingxiao Dong and Ragavan Srinivasan and Raj Ganapathy and Ramon Calderer and Ricardo Silveira Cabral and Robert Stojnic and Roberta Raileanu and Rohan Maheswari and Rohit Girdhar and Rohit Patel and Romain Sauvestre and Ronnie Polidoro and Roshan Sumbaly and Ross Taylor and Ruan Silva and Rui Hou and Rui Wang and Saghar Hosseini and Sahana Chennabasappa and Sanjay Singh and Sean Bell and Seohyun Sonia Kim and Sergey Edunov and Shaoliang Nie and Sharan Narang and Sharath Raparthy and Sheng Shen and Shengye Wan and Shruti Bhosale and Shun Zhang and Simon Vandenhende and Soumya Batra and Spencer Whitman and Sten Sootla and Stephane Collot and Suchin Gururangan and Sydney Borodinsky and Tamar Herman and Tara Fowler and Tarek Sheasha and Thomas Georgiou and Thomas Scialom and Tobias Speckbacher and Todor Mihaylov and Tong Xiao and Ujjwal Karn and Vedanuj Goswami and Vibhor Gupta and Vignesh Ramanathan and Viktor Kerkez and Vincent Gonguet and Virginie Do and Vish Vogeti and Vítor Albiero and Vladan Petrovic and Weiwei Chu and Wenhan Xiong and Wenyin Fu and Whitney Meers and Xavier Martinet and Xiaodong Wang and Xiaofang Wang and Xiaoqing Ellen Tan and Xide Xia and Xinfeng Xie and Xuchao Jia and Xuewei Wang and Yaelle Goldschlag and Yashesh Gaur and Yasmine Babaei and Yi Wen and Yiwen Song and Yuchen Zhang and Yue Li and Yuning Mao and Zacharie Delpierre Coudert and Zheng Yan and Zhengxing Chen and Zoe Papakipos and Aaditya Singh and Aayushi Srivastava and Abha Jain and Adam Kelsey and Adam Shajnfeld and Adithya Gangidi and Adolfo Victoria and Ahuva Goldstand and Ajay Menon and Ajay Sharma and Alex Boesenberg and Alexei Baevski and Allie Feinstein and Amanda Kallet and Amit Sangani and Amos Teo and Anam Yunus and Andrei Lupu and Andres Alvarado and Andrew Caples and Andrew Gu and Andrew Ho and Andrew Poulton and Andrew Ryan and Ankit Ramchandani and Annie Dong and Annie Franco and Anuj Goyal and Aparajita Saraf and Arkabandhu Chowdhury and Ashley Gabriel and Ashwin Bharambe and Assaf Eisenman and Azadeh Yazdan and Beau James and Ben Maurer and Benjamin Leonhardi and Bernie Huang and Beth Loyd and Beto De Paola and Bhargavi Paranjape and Bing Liu and Bo Wu and Boyu Ni and Braden Hancock and Bram Wasti and Brandon Spence and Brani Stojkovic and Brian Gamido and Britt Montalvo and Carl Parker and Carly Burton and Catalina Mejia and Ce Liu and Changhan Wang and Changkyu Kim and Chao Zhou and Chester Hu and Ching-Hsiang Chu and Chris Cai and Chris Tindal and Christoph Feichtenhofer and Cynthia Gao and Damon Civin and Dana Beaty and Daniel Kreymer and Daniel Li and David Adkins and David Xu and Davide Testuggine and Delia David and Devi Parikh and Diana Liskovich and Didem Foss and Dingkang Wang and Duc Le and Dustin Holland and Edward Dowling and Eissa Jamil and Elaine Montgomery and Eleonora Presani and Emily Hahn and Emily Wood and Eric-Tuan Le and Erik Brinkman and Esteban Arcaute and Evan Dunbar and Evan Smothers and Fei Sun and Felix Kreuk and Feng Tian and Filippos Kokkinos and Firat Ozgenel and Francesco Caggioni and Frank Kanayet and Frank Seide and Gabriela Medina Florez and Gabriella Schwarz and Gada Badeer and Georgia Swee and Gil Halpern and Grant Herman and Grigory Sizov and Guangyi and Zhang and Guna Lakshminarayanan and Hakan Inan and Hamid Shojanazeri and Han Zou and Hannah Wang and Hanwen Zha and Haroun Habeeb and Harrison Rudolph and Helen Suk and Henry Aspegren and Hunter Goldman and Hongyuan Zhan and Ibrahim Damlaj and Igor Molybog and Igor Tufanov and Ilias Leontiadis and Irina-Elena Veliche and Itai Gat and Jake Weissman and James Geboski and James Kohli and Janice Lam and Japhet Asher and Jean-Baptiste Gaya and Jeff Marcus and Jeff Tang and Jennifer Chan and Jenny Zhen and Jeremy Reizenstein and Jeremy Teboul and Jessica Zhong and Jian Jin and Jingyi Yang and Joe Cummings and Jon Carvill and Jon Shepard and Jonathan McPhie and Jonathan Torres and Josh Ginsburg and Junjie Wang and Kai Wu and Kam Hou U and Karan Saxena and Kartikay Khandelwal and Katayoun Zand and Kathy Matosich and Kaushik Veeraraghavan and Kelly Michelena and Keqian Li and Kiran Jagadeesh and Kun Huang and Kunal Chawla and Kyle Huang and Lailin Chen and Lakshya Garg and Lavender A and Leandro Silva and Lee Bell and Lei Zhang and Liangpeng Guo and Licheng Yu and Liron Moshkovich and Luca Wehrstedt and Madian Khabsa and Manav Avalani and Manish Bhatt and Martynas Mankus and Matan Hasson and Matthew Lennie and Matthias Reso and Maxim Groshev and Maxim Naumov and Maya Lathi and Meghan Keneally and Miao Liu and Michael L. Seltzer and Michal Valko and Michelle Restrepo and Mihir Patel and Mik Vyatskov and Mikayel Samvelyan and Mike Clark and Mike Macey and Mike Wang and Miquel Jubert Hermoso and Mo Metanat and Mohammad Rastegari and Munish Bansal and Nandhini Santhanam and Natascha Parks and Natasha White and Navyata Bawa and Nayan Singhal and Nick Egebo and Nicolas Usunier and Nikhil Mehta and Nikolay Pavlovich Laptev and Ning Dong and Norman Cheng and Oleg Chernoguz and Olivia Hart and Omkar Salpekar and Ozlem Kalinli and Parkin Kent and Parth Parekh and Paul Saab and Pavan Balaji and Pedro Rittner and Philip Bontrager and Pierre Roux and Piotr Dollar and Polina Zvyagina and Prashant Ratanchandani and Pritish Yuvraj and Qian Liang and Rachad Alao and Rachel Rodriguez and Rafi Ayub and Raghotham Murthy and Raghu Nayani and Rahul Mitra and Rangaprabhu Parthasarathy and Raymond Li and Rebekkah Hogan and Robin Battey and Rocky Wang and Russ Howes and Ruty Rinott and Sachin Mehta and Sachin Siby and Sai Jayesh Bondu and Samyak Datta and Sara Chugh and Sara Hunt and Sargun Dhillon and Sasha Sidorov and Satadru Pan and Saurabh Mahajan and Saurabh Verma and Seiji Yamamoto and Sharadh Ramaswamy and Shaun Lindsay and Shaun Lindsay and Sheng Feng and Shenghao Lin and Shengxin Cindy Zha and Shishir Patil and Shiva Shankar and Shuqiang Zhang and Shuqiang Zhang and Sinong Wang and Sneha Agarwal and Soji Sajuyigbe and Soumith Chintala and Stephanie Max and Stephen Chen and Steve Kehoe and Steve Satterfield and Sudarshan Govindaprasad and Sumit Gupta and Summer Deng and Sungmin Cho and Sunny Virk and Suraj Subramanian and Sy Choudhury and Sydney Goldman and Tal Remez and Tamar Glaser and Tamara Best and Thilo Koehler and Thomas Robinson and Tianhe Li and Tianjun Zhang and Tim Matthews and Timothy Chou and Tzook Shaked and Varun Vontimitta and Victoria Ajayi and Victoria Montanez and Vijai Mohan and Vinay Satish Kumar and Vishal Mangla and Vlad Ionescu and Vlad Poenaru and Vlad Tiberiu Mihailescu and Vladimir Ivanov and Wei Li and Wenchen Wang and Wenwen Jiang and Wes Bouaziz and Will Constable and Xiaocheng Tang and Xiaojian Wu and Xiaolan Wang and Xilun Wu and Xinbo Gao and Yaniv Kleinman and Yanjun Chen and Ye Hu and Ye Jia and Ye Qi and Yenda Li and Yilin Zhang and Ying Zhang and Yossi Adi and Youngjin Nam and Yu and Wang and Yu Zhao and Yuchen Hao and Yundi Qian and Yunlu Li and Yuzi He and Zach Rait and Zachary DeVito and Zef Rosnbrick and Zhaoduo Wen and Zhenyu Yang and Zhiwei Zhao and Zhiyu Ma},
      year={2024},
      eprint={2407.21783},
      archivePrefix={arXiv},
      primaryClass={cs.AI},
      url={https://arxiv.org/abs/2407.21783}, 
}

@misc{openai2024gpt4technicalreport,
      title={GPT-4 Technical Report}, 
      author={OpenAI and Josh Achiam and Steven Adler and Sandhini Agarwal and Lama Ahmad and Ilge Akkaya and Florencia Leoni Aleman and Diogo Almeida and Janko Altenschmidt and Sam Altman and Shyamal Anadkat and Red Avila and Igor Babuschkin and Suchir Balaji and Valerie Balcom and Paul Baltescu and Haiming Bao and Mohammad Bavarian and Jeff Belgum and Irwan Bello and Jake Berdine and Gabriel Bernadett-Shapiro and Christopher Berner and Lenny Bogdonoff and Oleg Boiko and Madelaine Boyd and Anna-Luisa Brakman and Greg Brockman and Tim Brooks and Miles Brundage and Kevin Button and Trevor Cai and Rosie Campbell and Andrew Cann and Brittany Carey and Chelsea Carlson and Rory Carmichael and Brooke Chan and Che Chang and Fotis Chantzis and Derek Chen and Sully Chen and Ruby Chen and Jason Chen and Mark Chen and Ben Chess and Chester Cho and Casey Chu and Hyung Won Chung and Dave Cummings and Jeremiah Currier and Yunxing Dai and Cory Decareaux and Thomas Degry and Noah Deutsch and Damien Deville and Arka Dhar and David Dohan and Steve Dowling and Sheila Dunning and Adrien Ecoffet and Atty Eleti and Tyna Eloundou and David Farhi and Liam Fedus and Niko Felix and Simón Posada Fishman and Juston Forte and Isabella Fulford and Leo Gao and Elie Georges and Christian Gibson and Vik Goel and Tarun Gogineni and Gabriel Goh and Rapha Gontijo-Lopes and Jonathan Gordon and Morgan Grafstein and Scott Gray and Ryan Greene and Joshua Gross and Shixiang Shane Gu and Yufei Guo and Chris Hallacy and Jesse Han and Jeff Harris and Yuchen He and Mike Heaton and Johannes Heidecke and Chris Hesse and Alan Hickey and Wade Hickey and Peter Hoeschele and Brandon Houghton and Kenny Hsu and Shengli Hu and Xin Hu and Joost Huizinga and Shantanu Jain and Shawn Jain and Joanne Jang and Angela Jiang and Roger Jiang and Haozhun Jin and Denny Jin and Shino Jomoto and Billie Jonn and Heewoo Jun and Tomer Kaftan and Łukasz Kaiser and Ali Kamali and Ingmar Kanitscheider and Nitish Shirish Keskar and Tabarak Khan and Logan Kilpatrick and Jong Wook Kim and Christina Kim and Yongjik Kim and Jan Hendrik Kirchner and Jamie Kiros and Matt Knight and Daniel Kokotajlo and Łukasz Kondraciuk and Andrew Kondrich and Aris Konstantinidis and Kyle Kosic and Gretchen Krueger and Vishal Kuo and Michael Lampe and Ikai Lan and Teddy Lee and Jan Leike and Jade Leung and Daniel Levy and Chak Ming Li and Rachel Lim and Molly Lin and Stephanie Lin and Mateusz Litwin and Theresa Lopez and Ryan Lowe and Patricia Lue and Anna Makanju and Kim Malfacini and Sam Manning and Todor Markov and Yaniv Markovski and Bianca Martin and Katie Mayer and Andrew Mayne and Bob McGrew and Scott Mayer McKinney and Christine McLeavey and Paul McMillan and Jake McNeil and David Medina and Aalok Mehta and Jacob Menick and Luke Metz and Andrey Mishchenko and Pamela Mishkin and Vinnie Monaco and Evan Morikawa and Daniel Mossing and Tong Mu and Mira Murati and Oleg Murk and David Mély and Ashvin Nair and Reiichiro Nakano and Rajeev Nayak and Arvind Neelakantan and Richard Ngo and Hyeonwoo Noh and Long Ouyang and Cullen O'Keefe and Jakub Pachocki and Alex Paino and Joe Palermo and Ashley Pantuliano and Giambattista Parascandolo and Joel Parish and Emy Parparita and Alex Passos and Mikhail Pavlov and Andrew Peng and Adam Perelman and Filipe de Avila Belbute Peres and Michael Petrov and Henrique Ponde de Oliveira Pinto and Michael and Pokorny and Michelle Pokrass and Vitchyr H. Pong and Tolly Powell and Alethea Power and Boris Power and Elizabeth Proehl and Raul Puri and Alec Radford and Jack Rae and Aditya Ramesh and Cameron Raymond and Francis Real and Kendra Rimbach and Carl Ross and Bob Rotsted and Henri Roussez and Nick Ryder and Mario Saltarelli and Ted Sanders and Shibani Santurkar and Girish Sastry and Heather Schmidt and David Schnurr and John Schulman and Daniel Selsam and Kyla Sheppard and Toki Sherbakov and Jessica Shieh and Sarah Shoker and Pranav Shyam and Szymon Sidor and Eric Sigler and Maddie Simens and Jordan Sitkin and Katarina Slama and Ian Sohl and Benjamin Sokolowsky and Yang Song and Natalie Staudacher and Felipe Petroski Such and Natalie Summers and Ilya Sutskever and Jie Tang and Nikolas Tezak and Madeleine B. Thompson and Phil Tillet and Amin Tootoonchian and Elizabeth Tseng and Preston Tuggle and Nick Turley and Jerry Tworek and Juan Felipe Cerón Uribe and Andrea Vallone and Arun Vijayvergiya and Chelsea Voss and Carroll Wainwright and Justin Jay Wang and Alvin Wang and Ben Wang and Jonathan Ward and Jason Wei and CJ Weinmann and Akila Welihinda and Peter Welinder and Jiayi Weng and Lilian Weng and Matt Wiethoff and Dave Willner and Clemens Winter and Samuel Wolrich and Hannah Wong and Lauren Workman and Sherwin Wu and Jeff Wu and Michael Wu and Kai Xiao and Tao Xu and Sarah Yoo and Kevin Yu and Qiming Yuan and Wojciech Zaremba and Rowan Zellers and Chong Zhang and Marvin Zhang and Shengjia Zhao and Tianhao Zheng and Juntang Zhuang and William Zhuk and Barret Zoph},
      year={2024},
      eprint={2303.08774},
      archivePrefix={arXiv},
      primaryClass={cs.CL},
      url={https://arxiv.org/abs/2303.08774}, 
}

@article{touvron2023llama,
  title={Llama 2: Open foundation and fine-tuned chat models},
  author={Touvron, Hugo and Martin, Louis and Stone, Kevin and Albert, Peter and Almahairi, Amjad and Babaei, Yasmine and Bashlykov, Nikolay and Batra, Soumya and Bhargava, Prajjwal and Bhosale, Shruti and others},
  journal={arXiv preprint arXiv:2307.09288},
  year={2023}
}

@article{rafailov2023direct,
  title={Direct preference optimization: Your language model is secretly a reward model},
  author={Rafailov, Rafael and Sharma, Archit and Mitchell, Eric and Manning, Christopher D and Ermon, Stefano and Finn, Chelsea},
  journal={Advances in neural information processing systems},
  volume={36},
  pages={53728--53741},
  year={2023}
}

@article{eldan2023s,
  title={Who’s harry potter? approximate unlearning for LLMs},
  author={Eldan, Ronen and Russinovich, Mark},
  year={2023}
}

@inproceedings{lin2004rouge,
  title={Rouge: A package for automatic evaluation of summaries},
  author={Lin, Chin-Yew},
  booktitle={Text summarization branches out},
  pages={74--81},
  year={2004}
}

@misc{zheng2026offsidebenchmarkingunlearningmisinformation,
      title={OFFSIDE: Benchmarking Unlearning Misinformation in Multimodal Large Language Models}, 
      author={Hao Zheng and Zirui Pang and Ling li and Zhijie Deng and Yuhan Pu and Zhaowei Zhu and Xiaobo Xia and Jiaheng Wei},
      year={2026},
      eprint={2510.22535},
      archivePrefix={arXiv},
      primaryClass={cs.AI},
      url={https://arxiv.org/abs/2510.22535}, 
}

@article{team2023gemini,
  title={Gemini: a family of highly capable multimodal models},
  author={Team, Gemini and Anil, Rohan and Borgeaud, Sebastian and Alayrac, Jean-Baptiste and Yu, Jiahui and Soricut, Radu and Schalkwyk, Johan and Dai, Andrew M and Hauth, Anja and Millican, Katie and others},
  journal={arXiv preprint arXiv:2312.11805},
  year={2023}
}

@article{guo2025deepseek,
  title={Deepseek-r1: Incentivizing reasoning capability in llms via reinforcement learning},
  author={Guo, Daya and Yang, Dejian and Zhang, Haowei and Song, Junxiao and Wang, Peiyi and Zhu, Qihao and Xu, Runxin and Zhang, Ruoyu and Ma, Shirong and Bi, Xiao and others},
  journal={arXiv preprint arXiv:2501.12948},
  year={2025}
}

@article{li2023textbooks,
  title={Textbooks are all you need ii: phi-1.5 technical report},
  author={Li, Yuanzhi and Bubeck, S{\'e}bastien and Eldan, Ronen and Del Giorno, Allie and Gunasekar, Suriya and Lee, Yin Tat},
  journal={arXiv preprint arXiv:2309.05463},
  year={2023}
}

@article{zhang2022opt,
  title={Opt: Open pre-trained transformer language models},
  author={Zhang, Susan and Roller, Stephen and Goyal, Naman and Artetxe, Mikel and Chen, Moya and Chen, Shuohui and Dewan, Christopher and Diab, Mona and Li, Xian and Lin, Xi Victoria and others},
  journal={arXiv preprint arXiv:2205.01068},
  year={2022}
}

@inproceedings{karamolegkou2023copyright,
  title={Copyright violations and large language models},
  author={Karamolegkou, Antonia and Li, Jiaang and Zhou, Li and S{\o}gaard, Anders},
  booktitle={Proceedings of the 2023 Conference on Empirical Methods in Natural Language Processing},
  pages={7403--7412},
  year={2023}
}

@article{grynbaum2023times,
  title={The Times sues OpenAI and Microsoft over AI use of copyrighted work},
  author={Grynbaum, Michael M and Mac, Ryan},
  journal={The New York Times},
  volume={27},
  number={1},
  year={2023}
}

@inproceedings{chu2024protect,
  title={How to protect copyright data in optimization of large language models?},
  author={Chu, Timothy and Song, Zhao and Yang, Chiwun},
  booktitle={Proceedings of the AAAI Conference on Artificial Intelligence},
  volume={38},
  number={16},
  pages={17871--17879},
  year={2024}
}

@article{zhang2024unlearncanvas,
  title={Unlearncanvas: Stylized image dataset for enhanced machine unlearning evaluation in diffusion models},
  author={Zhang, Yihua and Fan, Chongyu and Zhang, Yimeng and Yao, Yuguang and Jia, Jinghan and Liu, Jiancheng and Zhang, Gaoyuan and Liu, Gaowen and Kompella, Ramana Rao and Liu, Xiaoming and others},
  journal={arXiv preprint arXiv:2402.11846},
  year={2024}
}

@inproceedings{zhang2024generate,
  title={To generate or not? safety-driven unlearned diffusion models are still easy to generate unsafe images... for now},
  author={Zhang, Yimeng and Jia, Jinghan and Chen, Xin and Chen, Aochuan and Zhang, Yihua and Liu, Jiancheng and Ding, Ke and Liu, Sijia},
  booktitle={European Conference on Computer Vision},
  pages={385--403},
  year={2024},
  organization={Springer}
}

@inproceedings{staab2024beyond,
  title={Beyond memorization: Violating privacy via inference with large language models},
  author={Staab, Robin and Vero, Mark and Balunovic, Mislav and Vechev, Martin},
  booktitle={International Conference on Learning Representations},
  volume={2024},
  pages={33832--33878},
  year={2024}
}

@inproceedings{mireshghallah2024can,
  title={Can llms keep a secret? testing privacy implications of language models via contextual integrity theory},
  author={Mireshghallah, Niloofar and Kim, Hyunwoo and Zhou, Xuhui and Tsvetkov, Yulia and Sap, Maarten and Shokri, Reza and Choi, Yejin},
  booktitle={International Conference on Learning Representations},
  volume={2024},
  pages={1892--1915},
  year={2024}
}

@article{das2025security,
  title={Security and privacy challenges of large language models: A survey},
  author={Das, Badhan Chandra and Amini, M Hadi and Wu, Yanzhao},
  journal={ACM Computing Surveys},
  volume={57},
  number={6},
  pages={1--39},
  year={2025},
  publisher={ACM New York, NY}
}

@inproceedings{di2025adversarial,
  title={Adversarial machine unlearning},
  author={Di, Zonglin and Yu, Sixie and Vorobeychik, Yevgeniy and Liu, Yang},
  booktitle={International Conference on Learning Representations},
  volume={2025},
  pages={21612--21633},
  year={2025}
}

@misc{pu2026cameoconditionalqualityawaremultiagent,
      title={CAMEO: A Conditional and Quality-Aware Multi-Agent Image Editing Orchestrator}, 
      author={Yuhan Pu and Hao Zheng and Ziqian Mo and Hill Zhang and Tianyi Fan and Shuhong Wu and Jiaheng Wei},
      year={2026},
      eprint={2604.03156},
      archivePrefix={arXiv},
      primaryClass={cs.CV},
      url={https://arxiv.org/abs/2604.03156}, 
}

@misc{li2026ipbenchbenchmarkimageprotection,
      title={IP-Bench: Benchmark for Image Protection Methods in Image-to-Video Generation Scenarios}, 
      author={Xiaofeng Li and Leyi Sheng and Zhen Sun and Zongmin Zhang and Jiaheng Wei and Xinlei He},
      year={2026},
      eprint={2603.26154},
      archivePrefix={arXiv},
      primaryClass={cs.CV},
      url={https://arxiv.org/abs/2603.26154}, 
}

@misc{jia2026evianexplainablevisualinstructiontuning,
      title={Evian: Towards Explainable Visual Instruction-tuning Data Auditing}, 
      author={Zimu Jia and Mingjie Xu and Andrew Estornell and Jiaheng Wei},
      year={2026},
      eprint={2604.20544},
      archivePrefix={arXiv},
      primaryClass={cs.CV},
      url={https://arxiv.org/abs/2604.20544}, 
}

@misc{sun2026stegobattlefieldevaluatingimage,
      title={Stego Battlefield: Evaluating Image Steganography Attacks and Steganalysis Defenses}, 
      author={Zhen Sun and Zongmin Zhang and Leyi Sheng and Yule Liu and Yifan Liao and Ke Li and Xinhu Zheng and Jiaheng Wei and Wenyuan Yang and Xinlei He},
      year={2026},
      eprint={2605.05789},
      archivePrefix={arXiv},
      primaryClass={cs.CR},
      url={https://arxiv.org/abs/2605.05789}, 
}

@article{li2026recognition,
  title={Recognition through reasoning: Reinforcing image geo-localization with large vision-language models},
  author={Li, Ling and Zhou, Yao and Liang, Yuxuan and Tsung, Fugee and Wei, Jiaheng},
  journal={Advances in Neural Information Processing Systems},
  volume={38},
  pages={62132--62159},
  year={2026}
}

@article{deng2025lm,
  title={LM-mixup: Text Data Augmentation via Language Model based Mixup},
  author={Deng, Zhijie and Shen, Zhouan and Li, Ling and Zhou, Yao and Zhu, Zhaowei and He, Yanji and Wang, Wei and Wei, Jiaheng},
  journal={arXiv preprint arXiv:2510.20449},
  year={2025}
}

@article{yang2025entp,
  title={ENTP: Enhancing Low-Quality SFT Data via Neural-Symbolic Text Purge-Mix},
  author={Yang, Zile and Li, Ling and Di, Na and Pang, Jinlong and Zhou, Yao and Cheng, Hao and Han, Bo and Wei, Jiaheng},
  journal={arXiv preprint arXiv:2510.23160},
  year={2025}
}

@article{wu2026lifeside,
  title={LifeSide: Benchmarking Agents as Lifelong Digital Companions},
  author={Wu, Yuqian and Deng, Zhijie and Chen, Wei and Li, Junwei and Jiang, Yutian and Chen, Junle and Huang, Zhengjun and Liu, Qingxiang and Tang, Jing and Wei, Jiaheng and others},
  journal={arXiv preprint arXiv:2606.04660},
  year={2026}
}

@article{liu2025selectmix,
  title={SelectMix: Enhancing Label Noise Robustness through Targeted Sample Mixing},
  author={Liu, Qiuhao and Li, Ling and Lu, Yao and Xuan, Qi and Zhu, Zhaowei and Wei, Jiaheng},
  journal={arXiv preprint arXiv:2509.11265},
  year={2025}
}

@article{ji2024reversing,
  title={Reversing the forget-retain objectives: An efficient llm unlearning framework from logit difference},
  author={Ji, Jiabao and Liu, Yujian and Zhang, Yang and Liu, Gaowen and Kompella, Ramana R and Liu, Sijia and Chang, Shiyu},
  journal={Advances in Neural Information Processing Systems},
  volume={37},
  pages={12581--12611},
  year={2024}
}

@article{jia2024wagle,
  title={Wagle: Strategic weight attribution for effective and modular unlearning in large language models},
  author={Jia, Jinghan and Liu, Jiancheng and Zhang, Yihua and Ram, Parikshit and Baracaldo, Nathalie and Liu, Sijia},
  journal={Advances in Neural Information Processing Systems},
  volume={37},
  pages={55620--55646},
  year={2024}
}

@misc{wang2025invariancemakesllmunlearning,
      title={Invariance Makes LLM Unlearning Resilient Even to Unanticipated Downstream Fine-Tuning}, 
      author={Changsheng Wang and Yihua Zhang and Jinghan Jia and Parikshit Ram and Dennis Wei and Yuguang Yao and Soumyadeep Pal and Nathalie Baracaldo and Sijia Liu},
      year={2025},
      eprint={2506.01339},
      archivePrefix={arXiv},
      primaryClass={cs.LG},
      url={https://arxiv.org/abs/2506.01339}, 
}

@inproceedings{reisizadeh2026blur,
  title={Blur: A bi-level optimization approach for llm unlearning},
  author={Reisizadeh, Hadi and Jia, Jinghan and Bu, Zhiqi and Vinzamuri, Bhanukiran and Ramakrishna, Anil and Chang, Kai-Wei and Cevher, Volkan and Liu, Sijia and Hong, Mingyi},
  booktitle={Proceedings of the 19th Conference of the European Chapter of the Association for Computational Linguistics (Volume 1: Long Papers)},
  pages={7043--7058},
  year={2026}
}

@inproceedings{zhuang2025seuf,
  title={SEUF: Is Unlearning One Expert Enough for Mixture-of-Experts LLMs?},
  author={Zhuang, Haomin and Zhang, Yihua and Guo, Kehan and Jia, Jinghan and Liu, Gaowen and Liu, Sijia and Zhang, Xiangliang},
  booktitle={Proceedings of the 63rd Annual Meeting of the Association for Computational Linguistics (Volume 1: Long Papers)},
  pages={8664--8678},
  year={2025}
}

@article{fan2025towards,
  title={Towards llm unlearning resilient to relearning attacks: A sharpness-aware minimization perspective and beyond},
  author={Fan, Chongyu and Jia, Jinghan and Zhang, Yihua and Ramakrishna, Anil and Hong, Mingyi and Liu, Sijia},
  journal={arXiv preprint arXiv:2502.05374},
  year={2025}
}

\newpage
\appendix
The Appendix is organized as follows:

\begin{itemize}
    \item Appendix~\ref{app:data_generation} describes how we generate data for the three experiments.
    \item Appendix~\ref{app:baselines} introduces the baseline methods used in our experiments.
    \item Appendix~\ref{app:evaluation_metrics} presents the evaluation metrics used in our experiments.
    
    \item Appendix~\ref{app:exp_settings} describes the experimental settings and implementation details.
    \item Appendix~\ref{app:refusal_boundary} analyzes the generalisation advantage of RULE and \method over standard supervised fine-tuning.
    
    \item Appendix~\ref{app:ai_usage} provides a statement regarding the use of AI tools in this work.
\end{itemize}

\section{Data Generation}
\label{app:data_generation}

\subsection{Refusal Data Construction}

In the context of unlearning, we identify two essential types of queries that should be explicitly included in the refusal training set: \textbf{Type-I} queries that are likely to appear directly in the pretraining corpus (i.e., the forget set), and \textbf{Type-II} queries derived from them, such as QA-style questions that evaluate whether the model can still reason about the forgotten content. These two categories are particularly important because they capture the core knowledge that the model has either memorized or implicitly inferred from pretraining data.  In contrast, other semantically related variants (e.g., paraphrases, indirect references, or alternative phrasings) can typically be generalized through RL optimization. Therefore, we treat these two explicitly supervised categories as anchor cases that establish the model's refusal boundary, while RL further improves generalization to unseen but semantically related queries. Based on this principle, we construct the refusal data differently for each benchmark dataset.
\paragraph{RWKU.}
The dataset already provides QA-style queries (Type-II) used for rejection fine-tuning. We extend these queries via GPT-4o-mini to construct completion prompts, which aim to ask models to respond to the missing blank (Type-I). The construction prompt template is shown below:

\begin{tcolorbox}[
colback=gray!10,
colframe=gray!60,
title=Prompt for generating completion queries in RWKU,
fonttitle=\bfseries
]
\textcolor{blue}{[User]}

Transform the following question into a fill-in-the-blank declarative sentence. You may paraphrase the question to improve fluency. The sentence should be declarative and contain a blank represented by ``\_\_\_'', which does not have to appear at the end.

Original Question: \{query\}

\textcolor{red}{[Response]}
\end{tcolorbox}

\paragraph{MUSE-books.}
The dataset targets forgetting the ``Harry Potter'' book, which includes 3,045 raw text passages (Type-I). We construct QA-style queries (Type-II) directly from the source content. For each passage, we prompt GPT-4o-mini to generate three QA pairs, from which we randomly sample 841 final queries for training. We use the following QA construction prompt:

\begin{tcolorbox}[
colback=gray!10,
colframe=gray!60,
title=Prompt for generating QA queries in MUSE-books,
fonttitle=\bfseries
]

\textcolor{blue}{[User]}

Please generate three question-answer pairs based on the following context, the output format should be a json object:

\texttt{\{}\\
\texttt{"questions": [}\\
\texttt{\{}\\
\texttt{"question": "A single question related to the excerpt...",}\\
\texttt{"answer": "A precise answer extracted verbatim..."}\\
\texttt{\},}\\
\texttt{...}\\
\texttt{]}\\
\texttt{\}}

Input context: \{query\}

\textcolor{red}{[Response]}

\end{tcolorbox}

\paragraph{TOFU.}
The TOFU benchmark provides QA-style forget queries (Type-II) for fictional authors. Following the RULE setting, we further construct completion-style prompts (Type-I) from the original QA pairs. Specifically, we first identify the target author names from the forget set using rule-based name extraction. We then prompt GPT-4o-mini to transform each QA-style query into a fill-in-the-blank completion task that preserves the original semantic content while resembling pretraining-style text completion. Similar to RWKU, these completion prompts are used to improve the model's refusal behavior on pretraining-like memorization patterns. The resulting dataset therefore contains both QA-style refusal queries and completion-style refusal queries for Stage 1 rejection steering.

\begin{tcolorbox}[
colback=gray!10,
colframe=gray!60,
title=Prompt for generating completion queries in TOFU,
fonttitle=\bfseries
]

\textcolor{blue}{[User]}

Transform the following question-answer pair into a fill-in-the-blank declarative sentence. You may paraphrase the original content to improve fluency. The sentence should contain a blank represented by ``\_\_\_'' and should resemble a natural pretraining-style completion task.

Question: \{query\}

Answer: \{answer\}

\textcolor{red}{[Response]}

\end{tcolorbox}

\subsection{Boundary Data Construction}
\paragraph{Boundary Data.}
To construct boundary data for Stage 2 training, we adopt a controlled prompt rewriting strategy. Specifically, we prompt GPT-4o-mini to generate semantically similar neighbor queries by replacing the sensitive entity \(x\) in the forget prompt with a permissible counterpart \(x'\) (e.g., replacing ``Stephen King'' with ``J.K. Rowling''). The objective is to preserve the original writing style, semantic structure, and knowledge type while modifying the target entity. As a result, the generated boundary data remain highly similar to the forget queries in both semantics and format, but no longer belong to the removal target. We use the following prompt template for boundary data generation:

\begin{tcolorbox}[
colback=gray!10,
colframe=gray!60,
title=Prompt for generating neighbor queries,
fonttitle=\bfseries
]

\textcolor{blue}{[User]}

Rewrite the following question by replacing it with another well-known and real figure. Keep the writing style, sentence structure, and length as close as possible. Ensure that any referenced events or facts are real and accurate. Return the result in the following JSON format:

\texttt{\{}\\
\texttt{"question": "REWRITTEN\_QUESTION\_HERE",}\\
\texttt{"answer": "ACCURATE\_ANSWER\_HERE"}\\
\texttt{\}}

Original question:

\{question\}

\textcolor{red}{[Response]}

\end{tcolorbox}

\section{Baseline Methods}
\paragraph{Gradient Ascent (GA) \citep{yao2024large}.}
Gradient Ascent (GA) is a widely used baseline in machine unlearning. Instead of minimizing the training objective, GA directly maximizes the prediction loss on the forget dataset \(\mathcal{D}_f\). By encouraging the model to perform poorly on forget samples, the method pushes model parameters away from knowledge associated with \(\mathcal{D}_f\), thereby approximating the behavior of a model trained solely on the retain set \(\mathcal{D}_r\). The optimization objective is formulated as:

\begin{equation}
\label{eq:ga}
\mathcal{L}_{\text{GA}} ~=~
- \frac{1}{|\mathcal{D}_f|} \sum_{(x_i, y_i) \in \mathcal{D}_f} \ell(x_i, y_i;\theta).
\end{equation}

\paragraph{Gradient Difference (GD) \citep{liu2024large}.}
Gradient Difference (GD) is another commonly used baseline for machine unlearning. Different from GA, which only maximizes the loss on the forget set, GD jointly optimizes two opposing objectives: minimizing the prediction loss on the retain dataset \(\mathcal{D}_r\) while simultaneously maximizing the loss on the forget dataset \(\mathcal{D}_f\). This formulation aims to preserve useful knowledge from \(\mathcal{D}_r\) while removing information associated with \(\mathcal{D}_f\). The overall objective is defined as:

\begin{equation}
\label{eq:gd}
\begin{split}
\mathcal{L}_{\text{GD}}
~=~ \frac{1}{|\mathcal{D}_r|}
\sum_{(x_r, y_r)\in\mathcal{D}_r}
\ell(x_r, y_r;\theta) \\
- \frac{1}{|\mathcal{D}_f|}
\sum_{(x_f, y_f)\in\mathcal{D}_f}
\ell(x_f, y_f;\theta).
\end{split}
\end{equation}

\paragraph{KL Minimization (KL) \citep{maini2024tofu}.}
KL Minimization (KL) combines gradient ascent on the forget set \(\mathcal{D}_f\) with a KL-divergence regularization term on the retain set \(\mathcal{D}_r\). The method encourages forgetting by maximizing the prediction loss on forget samples, while simultaneously constraining the model outputs on retain samples to remain close to those of the original pretrained model \(M_{\hat{\theta}}\). This regularization helps preserve general utility during unlearning. The objective is formulated as:

\begin{equation}
\label{eq:kl}
\begin{split}
\mathcal{L}_{\text{KL}}
~=~
- \frac{1}{|\mathcal{D}_f|}
\sum_{(x_f,y_f)\in\mathcal{D}_f}
\ell(x_f,y_f;\theta) \\
+~
\frac{1}{|\mathcal{D}_r|}
\sum_{x_r\in\mathcal{D}_r}
\mathrm{KL}\!\left(
h_{\hat{\theta}}(x_r)\;\|\;h_{\theta}(x_r)
\right),
\end{split}
\end{equation}

where \(h_{\hat{\theta}}\) and \(h_\theta\) denote the output distributions produced by the original model and the unlearned model, respectively.

\paragraph{Preference Optimization (PO) \citep{maini2024tofu}.}
Preference Optimization (PO) modifies the model’s generation preference by encouraging refusal-style responses for prompts in the forget set \(\mathcal{D}_f\). To achieve this, an auxiliary dataset \(\mathcal{D}_{\text{IDK}}\) is constructed, where each forget prompt is paired with a safe refusal response such as ``I don’t know''. The training objective combines standard supervised fine-tuning on the retain set \(\mathcal{D}_r\) with refusal-oriented optimization on \(\mathcal{D}_{\text{IDK}}\). The objective is defined as:

\begin{equation}
\label{eq:po}
\begin{split}
\mathcal{L}_{\text{PO}}
~=~
\frac{1}{|\mathcal{D}_r|}
\sum_{(x_r,y_r)\in\mathcal{D}_r}
\ell(x_r,y_r;\theta) \\
+~
\frac{1}{|\mathcal{D}_{\text{IDK}}|}
\sum_{(x_f,y_{\text{IDK}})\in\mathcal{D}_{\text{IDK}}}
\ell(x_f,y_{\text{IDK}};\theta).
\end{split}
\end{equation}

This objective encourages the model to preserve its performance on the retain set while learning to refuse responses associated with the forget set.

\paragraph{Direct Preference Optimization (DPO) \citep{rafailov2023direct}.}
Direct Preference Optimization (DPO) extends preference optimization to the machine unlearning setting. Rather than optimizing preferences between human-chosen and rejected responses, DPO contrasts a safe refusal response \(y_e\) against the original forget response \(y_f\) under the same forget prompt \(x_f \in \mathcal{D}_f\). The objective encourages the model to assign higher preference to the refusal response, thereby suppressing knowledge associated with the forget set while maintaining overall model capability. Given inverse temperature parameter \(\beta\), the objective is defined as:

\begin{equation}
\label{eq:dpo}
\begin{split}
\mathcal{L}_{\text{DPO}}
~=~
-\,2\beta \;
\mathbb{E}_{x_f \in \mathcal{D}_f}
\Bigg[
\log \sigma\!\Big(
\beta \log h_\theta(x_f,y_e) \\
-~ \beta \log h_\theta(x_f,y_f)
- M_{\text{ref}}
\Big)
\Bigg],
\end{split}
\end{equation}

where \(h_\theta\) denotes the predictive distribution of the current model, and \(M_{\text{ref}}\) is an optional regularization term that penalizes deviation from the original pretrained model. A retention-aware variant additionally incorporates supervised learning on the retain set \(\mathcal{D}_r\) to better preserve general knowledge:

\begin{equation}
\label{eq:dpo-rt}
\mathcal{L}_{\text{DPO-RT}}
~=~
\mathcal{L}_{\text{DPO}}
\;+\;
\mathcal{L}(\mathcal{D}_r;\theta).
\end{equation}
\paragraph{Negative Preference Optimization (NPO) \citep{zhang2024negative}.}
Negative Preference Optimization (NPO) is designed to suppress undesired generations by directly penalizing responses associated with the forget set \(\mathcal{D}_f\). Unlike DPO, which relies on pairwise preference comparisons between preferred and dispreferred responses, NPO optimizes only the negative preference signal, explicitly discouraging the model from generating the original forget responses. Given inverse temperature parameter \(\beta\), the objective is formulated as:

\begin{equation}
\footnotesize
\label{eq:npo}
\begin{split}
\mathcal{L}_{\text{NPO}}
~=~
-\,2\beta \;
\mathbb{E}_{(x_f,y_f)\in\mathcal{D}_f}
\Big[
\log \sigma\!\big(
-\beta \log h_\theta(y_f|x_f)
\big)
\Big],
\end{split}
\end{equation}

where \(h_\theta\) denotes the predictive distribution of the current model. To better preserve model utility, a retention-aware variant further incorporates supervised optimization on the retain set \(\mathcal{D}_r\):

\begin{equation}
\label{eq:npo-rt}
\mathcal{L}_{\text{NPO-RT}}
~=~
\mathcal{L}_{\text{NPO}}
\;+\;
\mathcal{L}(\mathcal{D}_r;\theta).
\end{equation}

\paragraph{Mismatch.}
Mismatch extends the preference-based unlearning framework by introducing randomly constructed response sequences \citep{yao2024large}. Similar to PO, the method preserves model utility through supervised fine-tuning on the retain set \(\mathcal{D}_r\), while additionally optimizing a mismatch objective based on randomly sampled responses \(Y_{\text{rnd}}\). These random continuations are paired with forget prompts, encouraging the model to generate neutral or irrelevant outputs instead of memorized forget responses. The objective is formulated as:

\begin{equation}
\label{eq:mismatch}
\begin{split}
\mathcal{L}_{\text{Mismatch}}
~=~
\frac{1}{|\mathcal{D}_r|}
\sum_{(x_r,y_r)\in\mathcal{D}_r}
\ell(x_r,y_r;\theta) \\
+~
\frac{1}{|\mathcal{D}_{\text{rnd}}|}
\sum_{y_{\text{rnd}}\in Y_{\text{rnd}}}
\ell(x_f,y_{\text{rnd}};\theta),
\end{split}
\end{equation}

where \(Y_{\text{rnd}}\) denotes a set of randomly sampled responses associated with forget prompts \(x_f \in \mathcal{D}_f\).

\paragraph{LLMU \citep{yao2024large}.}LLMU extends Gradient Ascent by incorporating two auxiliary
components: (1) random-completion unlearning using sequences generated from
forget prompts, and (2) retention regularization on normal retain data. The
objective encourages forgetting by maximizing loss on the forget set
$\mathcal{D}_f$, while simultaneously training on random completions
$\mathcal{D}_{\text{rand}}$ and aligning the model’s predictions on the retain
set $\mathcal{D}_{\text{normal}}$ with the original model through KL divergence.
Formally, the loss is:
\begin{equation}
\label{eq:llmu}
\begin{aligned}
\mathcal{L}_{\text{LLMU}}
~=~ &-\,\frac{\varepsilon_1}{|\mathcal{D}_f|}
\sum_{(x_f,y_f)\in\mathcal{D}_f}\ell(x_f,y_f;\theta) \\
&+~ \frac{\varepsilon_2}{|\mathcal{D}_{\text{rand}}|}
\sum_{x\in\mathcal{D}_{\text{rand}}}\ell(x;\theta) \\
&+~ \frac{\varepsilon_3}{|\mathcal{D}_{\text{normal}}|}
\sum_{x\in\mathcal{D}_{\text{normal}}}
\mathrm{KL}\!\left(h_{\hat{\theta}}(x)\;\|\;h_\theta(x)\right),
\end{aligned}
\end{equation}

\paragraph{Task Vectors \citep{eldan2023s}.}
Task Vectors construct an unlearned model by explicitly reversing the parameter update direction associated with the forget set \(\mathcal{D}_f\). Let \(\theta_o\) denote the parameters of the original pretrained model, and let \(\theta_{\text{reinforce}}\) represent the parameters of a model additionally fine-tuned to strongly memorize the forget data. The difference between these two parameter sets defines the task vector \((\theta_{\text{reinforce}} - \theta_o)\). The unlearning process is then performed by subtracting this vector from the original model parameters:

\begin{equation}
\label{eq:taskvector}
\theta ~=~ \theta_o - (\theta_{\text{reinforce}} - \theta_o).
\end{equation}

This approach removes knowledge associated with \(\mathcal{D}_f\) by moving the model parameters in the opposite direction of the forget-data adaptation, without requiring additional optimization steps.

\paragraph{Who’s Harry Potter (WHP) \citep{eldan2023s}.}
Who’s Harry Potter (WHP) performs unlearning through distribution-level interpolation between the original model \(\theta_o\) and a reinforced model \(\theta_{\text{reinforce}}\) that is intentionally overfitted on the forget set. Let \(p_\theta(\cdot|x)\) denote the token-level output distribution of the unlearned model for input \(x\). WHP modifies this distribution by subtracting a scaled adaptation direction derived from the reinforced model:

\begin{equation}
\footnotesize
\label{eq:whp}
p_\theta(\cdot|x)
~=~ p_{\theta_o}(\cdot|x) - \alpha \big(p_{\theta_{\text{reinforce}}}(\cdot|x)
- p_{\theta_o}(\cdot|x)\big),
\end{equation}

where \(\alpha\) controls the strength of the forgetting operation. This formulation steers the model distribution away from the forget-oriented reinforced model while maintaining proximity to the original pretrained model distribution.

\paragraph{FLAT \citep{wang2024llm}.}
Forget data only Loss AdjustmenT (FLAT) is a loss-adjustment-based unlearning method that does not require retain data or an additional reference model. Instead of directly applying gradient ascent on the forget set \(\mathcal{D}_f\), FLAT optimizes an \(f\)-divergence objective between a safe template response \(y_e\) (such as a refusal or irrelevant completion) and the original forget response \(y_f\). For each forget sample \((x_f, y_f)\), the method introduces a paired safe response and encourages the model to move toward the safe template distribution while suppressing the original forget response. The objective is defined as:

\begin{equation}
\label{eq:flat}
\begin{split}
\mathcal{L}_{\text{FLAT}}
~=~
-\,g^{\ast}\!\left(
P(x_f,y_e;\theta)
\right) \\
+~
f^{\ast}\!\left(
g^{\ast}\!\left(
P(x_f,y_f;\theta)
\right)
\right),
\end{split}
\end{equation}

where \(P(x,y;\theta)\) denotes the average token-level prediction probability of response \(y\) conditioned on input \(x\), while \(g^{\ast}(\cdot)\) and \(f^{\ast}(\cdot)\) represent the variational and conjugate functions associated with the selected \(f\)-divergence. This formulation enables effective forgetting by encouraging alignment with safe responses instead of directly optimizing against forget samples.
\paragraph{SGA \citep{pang2025labelsmoothingimprovesgradient}.}Smooth Gradient Ascent (SGA) adopts a Generalized Label Smoothing paradigm that jointly trains on forget data and model-generated normal data. By incorporating normal data during optimization, SGA alleviates the gradient collapse issue commonly observed in traditional Gradient Ascent methods. Specifically, the model simultaneously learns to forget target knowledge while maintaining stable behavior on normal generations. The optimization objective is formulated as:

\begin{equation}
\label{eq:sga}
\begin{split}
\min_{\theta} \;
\underbrace{
\left(1-r+\tfrac{r}{K}\right)\,
\mathbb{E}_{(x_f, y_f) \in \mathcal{D}_f} 
\big[
\ell_f(y_f \mid x_f; \theta)
\big]
}_{\text{forget}}
\\
+~
\underbrace{
\left(\tfrac{r}{K}\right)\,
\mathbb{E}_{(x_p, y_p) \in \mathcal{D}_p} 
\Bigg[
\sum_{k=1}^{K}
\ell_p^{(k)}
(y_p^{(k)} \mid x_p^{(k)}; \theta)
\Bigg]
}_{\text{normal data}}.
\end{split}
\end{equation}

\paragraph{SimNPO \citep{fan2025simplicityprevailsrethinkingnegative}.}
SimNPO is a simplified variant of NPO that removes the dependency on a reference model and introduces length normalization for more stable optimization. The objective directly suppresses the likelihood of forget responses while normalizing over response length. The objective is defined as:

\begin{equation}
\footnotesize
\label{eq:simnpo}
\begin{split}
\mathcal{L}_{\text{SimNPO}}
~=~
-\,2\beta \;
\mathbb{E}_{(x_f,y_f)\in\mathcal{D}_f}
\\
\Bigg[
\log \sigma \Big(
- \frac{\beta}{|y_f|}
\log h_\theta(y_f|x_f)
- \gamma
\Big)
\Bigg].
\end{split}
\end{equation}
\label{app:baselines}

\section{Evaluation Metrics}
\label{app:evaluation_metrics}
\subsection{RWKU}

\paragraph{Forget Quality.}
Forget Quality evaluates whether the model has successfully removed target knowledge after unlearning. In RWKU, this metric is evaluated on three categories of forget tasks: Fact-Based (FB), Question-Answering (QA), and Author Attribution (AA). FB measures whether the model can still recall factual information related to the forget set, QA evaluates the model’s ability to answer forget-related questions, and AA tests whether the model can correctly identify authorship or source-related information associated with forgotten content. Lower scores indicate better forgetting performance.

\paragraph{Retain Quality.}
Retain Quality measures whether the model preserves useful knowledge and general capabilities after unlearning. In RWKU, this metric is evaluated on retain-set tasks including Fact-Based (FB) and Question-Answering (QA) evaluations. Higher scores indicate better preservation of general utility and retained knowledge.
\paragraph{Forget Naturalness.}
In RWKU, Forget Naturalness evaluates whether the unlearned model still maintains normal language quality and general utility after forgetting. The metric mainly measures the fluency, truthfulness, and downstream capability of the model on non-forget tasks, reflecting whether the model still behaves like a standard aligned LLM after unlearning.

\paragraph{Tokens.}This refers to the proportion of data used in the RWKU task compared with the full forget and retain data. For Stage 2 of RULE and \method, since the training is conducted on the constructed boundary set, the token ratio is smaller than 100\%.

\subsection{MUSE}
\paragraph{No Verbatim Memorization (VerbMem).}
To assess whether the model has completely unlearned the target content, we evaluate verbatim memorization (\emph{VerbMem}). This metric measures the similarity between the model’s continuation and the ground-truth continuation from the forget set, restricted to the first $l$ tokens of each sample. Following prior work, we use the ROUGE-L F1 \citep{lin2004rouge} score as the evaluation metric:
\begin{equation}
\footnotesize
\mathrm{VerbMem}(f, \mathcal{D}_{f})
= \frac{1}{|\mathcal{D}_{f}|} \sum_{x \in \mathcal{D}_{f}}
\mathrm{ROUGE}\big(f(x_{[:l]}),\, x_{[l+1:]} \big).
\end{equation}

\paragraph{No Knowledge Memorization (KnowMem).}
Knowledge memorization (\emph{KnowMem}) measures whether the model retains factual information about forgotten records. For each question–answer pair $(q, a)$ in the forget set $\mathcal{D}_{f}$, we compute the ROUGE score between the model’s predicted answer $f(q)$ and the ground-truth answer $a$, and then average across all samples:
\begin{equation}
\footnotesize
\mathrm{KnowMem}(f, \mathcal{D}_{f})
= \frac{1}{|\mathcal{D}_{f}|} \sum_{(q,a) \in \mathcal{D}_{f}}
\mathrm{ROUGE}\big(f(q),\, a \big).
\end{equation}


\paragraph{Utility Preservation.}
Finally, we evaluate whether the model preserves its general capabilities after unlearning. This is measured on the retain set $\mathcal{D}_{r}$ by computing the knowledge memorization score:
\begin{equation}
\mathrm{UtilityPreservation}
= \mathrm{KnowMem}(f_{\text{unlearn}}, \mathcal{D}_{r}).
\end{equation}

\paragraph{Forget Naturalness.}
In MUSE, Forget Naturalness mainly evaluates the quality of refusal responses generated for forget queries. The metric focuses on whether the model can produce natural, fluent, and assistant-like refusals while avoiding memorized target knowledge. Specifically, it considers factors such as readability, helpfulness, and truthfulness of the generated responses.

\paragraph{Tokens.}Same as Tokens in RWKU's metircs.
\subsection{TOFU}
\paragraph{Probability.} 
For each instance in either the retain or forget set, we compute the normalized conditional probability\[
P(a \mid q)^{1/|a|},
\]where $q$ denotes the input question, $a$ is a candidate answer, and $|a|$ is the token length of $a$. 
In the \textit{real authors} and \textit{world facts} subsets, the dataset provides five candidate answers 
$\{a_0, \tilde{a}_1, \tilde{a}_2, \tilde{a}_3, \tilde{a}_4\}$, where $a_0$ is the correct answer and each $\tilde{a}_i$ is a perturbed (incorrect) alternative. 
The probability ratio is defined as:
\[
\text{Probability} = \frac{P(a_0 \mid q)^{1/|a_0|}}{\sum_{i=1}^4 P(\tilde{a}_i \mid q)^{1/|\tilde{a}_i|}}.
\tag{25}
\]

\paragraph{Truth Ratio.} 
The truth ratio quantifies the model’s preference for perturbed answers. 
It is computed as the geometric mean of the normalized probabilities of all perturbed answers $\{\tilde{a}_1, \tilde{a}_2, \ldots\}$ relative to the normalized probability of the paraphrased answer $\hat{a}$:
\[
R_{\text{truth}} = \frac{\Big(\prod_{i=1}^{|A|} P(\tilde{a}_i \mid q)^{1/|\tilde{a}_i|}\Big)^{1/|A|}}{P(\hat{a} \mid q)^{1/|\hat{a}|}}.
\tag{26}
\]
In the \textit{real authors} and \textit{world facts} subsets, where paraphrased answers are not available, the original answer $a$ is used in the denominator. 

\paragraph{ROUGE-L.} 
For all TOFU subsets, we report the ROUGE-L recall score \citep{lin2004rouge} between ground-truth answers in the forget set and the model outputs after unlearning.

\paragraph{Model Utility.} 
Model utility is defined as the harmonic mean of nine scores, covering answer probability, truth ratio, and ROUGE-L recall across the retain, real authors, and world facts subsets. 
A higher utility score reflects stronger overall performance. 

\paragraph{Forget Quality.} 
Forget quality is evaluated using a Kolmogorov--Smirnov (KS) test that compares the distributions of truth ratios between the retained and unlearned models on the forget set. 
A higher $p$-value supports the null hypothesis that the two distributions are statistically indistinguishable, indicating consistent behavior between the retained and unlearned models.

\section{Experimental Settings}
\label{app:exp_settings}
All experiments are conducted on four NVIDIA A800 GPUs. For each experiment:
\paragraph{RWKU Setup.}
All RWKU experiments use a rollout batch size of 32 and a global actor batch size of 32. The micro-batch size per device is set to 8 for parameter updates and 16 for experience generation. We use a learning rate of 2e-6, a KL coefficient of 1e-2, and train the model for a maximum of 37 steps. During GRPO sampling, 8 rollouts are generated for each prompt.

For hybrid training, the replay buffer is activated after at least 100 low-reward groups are collected and the training step exceeds 15. Groups with mean reward lower than 0.6 are stored in the replay buffer. The maximum replay buffer size is set to 1200 groups, and each off-policy update samples 32 groups from the replay buffer.

\paragraph{MUSE Setup.}
All MUSE experiments use a rollout batch size of 32 and a global actor batch size of 32. The micro-batch size per device is set to 8 for parameter updates and 16 for experience generation. We use a learning rate of 2e-6 and a KL coefficient of 1e-2. During GRPO sampling, 8 rollouts are generated for each prompt.

For hybrid training, the replay buffer is activated after at least 500 low-reward groups are collected and the training step exceeds 26. Groups with mean reward lower than 0.4 are stored in the replay buffer. The maximum replay buffer size is set to 1200 groups, and each off-policy update samples 32 groups from the replay buffer.

\paragraph{TOFU Setup.}
All TOFU experiments use a rollout batch size of 32 and a global actor batch size of 32. The micro-batch size per device is set to 4 for parameter updates and 8 for experience generation. We use a learning rate of 2e-6 and a KL coefficient of 1e-2. During GRPO sampling, 8 rollouts are generated for each prompt. The maximum response length is set to 256, and the model is trained for 11 steps.

For hybrid training, the replay buffer is activated after at least 80 low-reward groups are collected and the training step exceeds 6. Groups with mean reward lower than 0.6 are stored in the replay buffer. The maximum replay buffer size is set to 400 groups, and each off-policy update samples 32 groups from the replay buffer.

\section{Theoretical Analysis: Generalisation Advantage of RULE and \method}
\label{app:refusal_boundary}

\textbf{Theorem 1}
\textit{(Generalisation Advantage of RULE and \method).}
Let $\Pi$ be a policy class with token-wise Rademacher complexity $C(\Pi)$
on sequences of length $H$. Define the mis-refusal risk as:
\begin{equation}
\label{eq:mis_refusal_risk}
\begin{aligned}
\mathcal{R}(\pi)
=
&\underbrace{
\Pr_{x\sim P_f^*}
\left[
\pi(x)\neq [\mathrm{refuse}]
\right]
}_{\text{(i) miss-refusal on forget}}
\\
&+
\underbrace{
\Pr_{x\sim P_r}
\left[
\pi(x)= [\mathrm{refuse}]
\right]
}_{\text{(ii) false-refusal on retain}} .
\end{aligned}
\end{equation}

\paragraph{(a) SFT.}
Empirical risk minimisation over a forget set $D_f$ of size $n_f$, using a
bounded loss $\ell\in[0,1]$, yields:
\begin{equation}
\label{eq:sft_bound}
\mathbb{E}
\left[
\mathcal{R}(\hat{\pi}_{\mathrm{sft}})
\right]
\leq
2
\sqrt{
\frac{
C(\Pi)
}{
n_f
}
}
+
\Delta_f
+
\underbrace{1}_{\Delta_r},
\end{equation}
where
\begin{equation*}
\Delta_f
=
\Pr_{x\sim P_f^*\setminus D_f}
[\cdot]
\end{equation*}
is the coverage gap on the forget set, and the final term represents
worst-case retain-side risk due to no supervision on retain-side queries.

\paragraph{(b) RULE.}
After $K$ on-policy updates collecting $m$ boundary prompts and $H$-length
rollouts per prompt, the returned policy $\hat{\pi}_{\mathrm{rule}}$ satisfies,
with probability at least $1-\delta$:
\begin{equation}
\label{eq:rule_bound}
\begin{aligned}
\mathcal{R}
\left(
\hat{\pi}_{\mathrm{rule}}
\right)
\leq
&\;
2
\sqrt{
\frac{
C(\Pi)
}{
n_f
+
KmH
}
}
+
\Delta_f
\\
&+
\epsilon_{\mathrm{EXPLORE}}
(K,m,H,\delta).
\end{aligned}
\end{equation}
where the exploration error is bounded as
\begin{equation*}
\label{eq:rule_exploration_error}
\epsilon_{\mathrm{EXPLORE}}
=
O
\left(
\sqrt{
\frac{
\log(1/\delta)
}{
KmH
}
}
\right).
\end{equation*}
Hence, for equal token budget
\begin{equation*}
n_f
\approx
KmH,
\end{equation*}
and under mild exploration, i.e.,
\begin{equation*}
\epsilon_{\mathrm{EXPLORE}}
<
1,
\end{equation*}
we obtain:
\begin{equation}
\label{eq:rule_better_than_sft}
\boxed{
\mathbb{E}
\left[
\mathcal{R}
\left(
\hat{\pi}_{\mathrm{rule}}
\right)
\right]
<
\mathbb{E}
\left[
\mathcal{R}
\left(
\hat{\pi}_{\mathrm{sft}}
\right)
\right]
}
\end{equation}
That is, RULE improves the worst-case refusal performance compared to SFT.

\paragraph{(c) \method.}
\method further augments RULE with a hard-case reply buffer. Given a reward
threshold $\tau$, we divide prompts into Easy and Hard regions:
\begin{equation*}
\label{eq:easy_region}
\mathcal{E}
=
\left\{
x:
\mathbb{E}_{y\sim\pi_\theta(\cdot|x)}
\left[
r(x,y)
\right]
\geq
\tau
\right\},
\end{equation*}
and
\begin{equation*}
\label{eq:hard_region}
\mathcal{H}
=
\left\{
x:
\mathbb{E}_{y\sim\pi_\theta(\cdot|x)}
\left[
r(x,y)
\right]
<
\tau
\right\}.
\end{equation*}
Since hard cases may appear on both the forget side and the retain/boundary
side, the risk can be further decomposed as:
\begin{equation}
\label{eq:risk_decomposition}
\begin{aligned}
\mathcal{R}(\pi)
=
&
p_f(\mathcal{E})
\mathcal{R}_{f,\mathcal{E}}(\pi)
+
p_f(\mathcal{H})
\mathcal{R}_{f,\mathcal{H}}(\pi)
\\
&
+
p_r(\mathcal{E})
\mathcal{R}_{r,\mathcal{E}}(\pi)
+
p_r(\mathcal{H})
\mathcal{R}_{r,\mathcal{H}}(\pi).
\end{aligned}
\end{equation}
Here, $\mathcal{R}_{f,\mathcal{H}}$ denotes the miss-refusal risk on hard
forget prompts, while $\mathcal{R}_{r,\mathcal{H}}$ denotes the false-refusal
risk on hard retain or boundary prompts.

For pure on-policy RULE, the expected number of hard-region rollout tokens is:
\begin{equation*}
\label{eq:on_policy_hard_tokens}
N_{\mathcal{H}}^{\mathrm{on}}
=
KmH
\cdot
p(\mathcal{H}).
\end{equation*}
Thus, the hard-region exploration error of pure RULE can be written as:
\begin{equation}
\label{eq:rule_hard_exploration_error}
\epsilon_{\mathcal{H}}^{\mathrm{RULE}}
=
O
\left(
\sqrt{
\frac{
\log(1/\delta)
}{
N_{\mathcal{H}}^{\mathrm{on}}
}
}
\right).
\end{equation}

In contrast, \method stores rollout groups whose mean reward is lower than
$\tau$ into a reply buffer $\mathcal{B}$ and reuses them during hybrid
training. Suppose the hybrid stage performs $L$ off-policy replay updates,
each using $b$ hard groups with $H$-length rollouts. The number of replayed
hard-region rollout tokens is:
\begin{equation*}
\label{eq:off_policy_hard_tokens}
N_{\mathcal{H}}^{\mathrm{off}}
=
LbH.
\end{equation*}

Since the replayed trajectories are generated by previous policies, \method
uses a clipped and normalized importance ratio:
\begin{equation*}
\label{eq:importance_ratio}
\tilde{\rho}(y|x)
=
\operatorname{Normalize}
\left(
\operatorname{Clip}
\left(
\frac{
\pi_\theta(y|x)
}{
\pi_{\theta_{\mathrm{old}}}(y|x)
}
\right)
\right).
\end{equation*}
Let $\eta_\rho\in(0,1]$ denote the effective-sample-size coefficient induced
by importance weighting:
\begin{equation*}
\label{eq:eta_rho}
\eta_\rho
=
\frac{
\left(
\mathbb{E}
\left[
\tilde{\rho}
\right]
\right)^2
}{
\mathbb{E}
\left[
\tilde{\rho}^2
\right]
}.
\end{equation*}
Then the effective number of hard-region training tokens in \method becomes:
\begin{equation*}
\label{eq:rulepp_effective_hard_tokens}
N_{\mathcal{H}}^{\method}
=
N_{\mathcal{H}}^{\mathrm{on}}
+
\eta_\rho
N_{\mathcal{H}}^{\mathrm{off}}.
\end{equation*}
Therefore, the hard-region exploration error of \method is bounded as:
\begin{equation}
\label{eq:rulepp_hard_exploration_error}
\epsilon_{\mathcal{H}}^{\method}
=
O
\left(
\sqrt{
\frac{
\log(1/\delta)
}{
N_{\mathcal{H}}^{\mathrm{on}}
+
\eta_\rho
N_{\mathcal{H}}^{\mathrm{off}}
}
}
\right).
\end{equation}

\begin{equation*}
\label{eq:rulepp_effective_sample_size}
N_{\mathcal{H}}^{\method}
=
N_{\mathcal{H}}^{\mathrm{on}}
+
\eta_\rho
N_{\mathcal{H}}^{\mathrm{off}}.
\end{equation*}

Since
\begin{equation*}
\label{eq:hard_token_comparison}
N_{\mathcal{H}}^{\method}
\geq
N_{\mathcal{H}}^{\mathrm{on}},
\end{equation*}
we have:
\begin{equation*}
\label{eq:hard_error_comparison}
\epsilon_{\mathcal{H}}^{\method}
\leq
\epsilon_{\mathcal{H}}^{\mathrm{RULE}}.
\end{equation*}

Combining the standard uniform convergence term with the hard-region replay
benefit, the policy returned by \method satisfies, with probability at least
$1-\delta$:
\begin{equation}
\label{eq:rulepp_bound}
\begin{aligned}
\mathcal{R}
\left(
\hat{\pi}_{\method}
\right)
\leq
&\;
2
\sqrt{
\frac{
C(\Pi)
}{
n_f
+
KmH
+
\eta_\rho
N_{\mathcal{H}}^{\mathrm{off}}
}
}
\\
&+
\Delta_f
+
\epsilon_{\mathcal{H}}^{\method}
+
\epsilon_{\mathrm{IS}}.
\end{aligned}
\end{equation}
where $\epsilon_{\mathrm{IS}}$ denotes the bias introduced by clipped
importance sampling.

Hence, when the clipping bias is controlled and the reduction of hard-region
exploration error dominates this bias, we obtain:
\begin{equation}
\label{eq:rulepp_better_than_rule}
\boxed{
\mathcal{R}
\left(
\hat{\pi}_{\method}
\right)
<
\mathcal{R}
\left(
\hat{\pi}_{\mathrm{rule}}
\right)
}
\end{equation}
That is, \method further improves RULE by reallocating effective training
tokens toward hard cases, reducing inefficient repeated sampling on Easy
cases and yielding a tighter hard-region generalisation bound.

\section{AI Usage Statement}
\label{app:ai_usage}
This paper only uses AI tools for language polishing and writing refinement.

\end{document}